\documentclass{article} % For LaTeX2e
\usepackage{iclr2026_conference,times}
\iclrfinalcopy

\usepackage{amsmath,amsfonts,bm}

\def\eqref#1{equation~\ref{#1}}
\def\1{\bm{1}}

\DeclareMathAlphabet{\mathsfit}{\encodingdefault}{\sfdefault}{m}{sl}
\SetMathAlphabet{\mathsfit}{bold}{\encodingdefault}{\sfdefault}{bx}{n}

\usepackage{hyperref}
\usepackage{url}
\usepackage{graphicx} 
\usepackage{wrapfig}

\newtheorem{Theorem}{Theorem}
\newtheorem{Lemma}{Lemma}

\usepackage[table]{xcolor}
\usepackage{booktabs}
\usepackage{adjustbox} 
\newcolumntype{g}{>{\columncolor{gray!12}}c}
\newcolumntype{w}{>{\columncolor{white}}c}
\usepackage{enumitem}
\usepackage{makecell}

\title{HiLoRA: Adaptive Hierarchical LoRA Routing for Training-Free Domain Generalization}

\author{
	Ziyi Han$^1$, Huanyu Wang$^2$, Zeyu Zhang$^1$, Xiangxiang Dai$^1$,  Xutong Liu$^3$, John C.S. Lui$^1$\\
	$^1$The Chinese University of Hong Kong, Hong Kong\\
	$^2$FSI Lab, Huawei Technologies Co., Ltd.\\
	$^3$Carnegie Mellon University, Pittsburgh, PA, USA\\
    zyhan24@cse.cuhk.edu.hk
}
\date{} 

\begin{document}

\maketitle

\begin{abstract}
    Low-Rank Adaptation (LoRA) has emerged as a widely used technique for adapting large language models (LLMs) to new domains, due to its modular design and broad availability on platforms such as HuggingFace. This availability has motivated efforts to reuse existing LoRAs for domain generalization. 
    However, existing methods often rely on explicit task labels or additional training, which are impractical for deployment. Moreover, they typically activate a fixed number of entire LoRA modules, leading to parameter redundancy or insufficiency that degrade performance.
    In this paper, we propose \texttt{HiLoRA}, a training-free framework that performs adaptive hierarchical routing over LoRA pools. Drawing on structural properties of LoRA, we define rank-one components (ROCs), in which each rank parameter is regarded as an independent unit. For a given input sequence, \texttt{HiLoRA} first adaptively selects a subset of LoRAs and determines their ROC allocation based on Gaussian likelihoods at the sequence level. At the token level, it further refines routing by activating only the most informative ROCs.
    We further provide theoretical guarantees that \texttt{HiLoRA} selects the most relevant LoRAs with high probability.
    Extensive experiments show that \texttt{HiLoRA} achieves substantial improvements in domain generalization, with accuracy gains of up to {\small $55\%$} over state-of-the-art baselines, while maintaining comparable inference throughput.%\footnote{Code will be released at: [link]}
    
\end{abstract}

\section{Introduction}\label{sec_intro}
Large Language Models (LLMs) have demonstrated remarkable capabilities across a wide variety of tasks \citep{zhou2024survey,naveed2025comprehensive}. 
However, adapting LLMs to specialized domains or tasks requires computationally expensive full fine-tuning \citep{hu2022lora}. 
To mitigate this cost, parameter-efficient fine-tuning (PEFT) techniques have been developed~\citep{ding2023parameter}. 
Among them, Low-Rank Adaptation (LoRA)~\citep{hu2022lora, tian2024hydralora} has become one of the most effective and widely adopted methods.
LoRA introduces lightweight low-rank matrices into selected layers of an LLM, thereby substantially reducing the number of trainable parameters while preserving strong downstream task performance.
Building on this success, community platforms such as HuggingFace\citep{huggingface} and ModelScope~\citep{modelscope} now host thousands of task-specific LoRA modules trained across diverse domains.
This rapidly expanding repository creates a unique opportunity: instead of training a new model for every task, one can directly exploit existing LoRAs to achieve scalable multi-domain adaptation.

However, realizing this potential is highly non-trivial, as effectively utilizing community-shared LoRAs introduces several challenges. 
\textit{First}, explicit task labels of inputs are typically unavailable in practice. If such labels were known, inputs from seen tasks could be directly routed to their specialized LoRAs, while unseen tasks could be aligned with related LoRAs based on task similarity. 
Without labels, however, distinguishing between seen and unseen cases and assigning appropriate LoRAs becomes highly challenging.
\textit{Second}, For a given input, activating too many LoRAs or entire modules leads to parameter redundancy and interference, whereas activating too few may discard valuable knowledge, ultimately reducing accuracy~\citep{cheng2025sci}.
\textit{Third}, as repositories continue to expand with thousands of task-specific LoRAs, the routing mechanism must remain computationally efficient to ensure scalability \citep{ostapenko2024towards}.

\begin{wrapfigure}{r}{0.35\linewidth}
    %\vspace{-3mm}
    \centering
    \includegraphics[width=\linewidth]{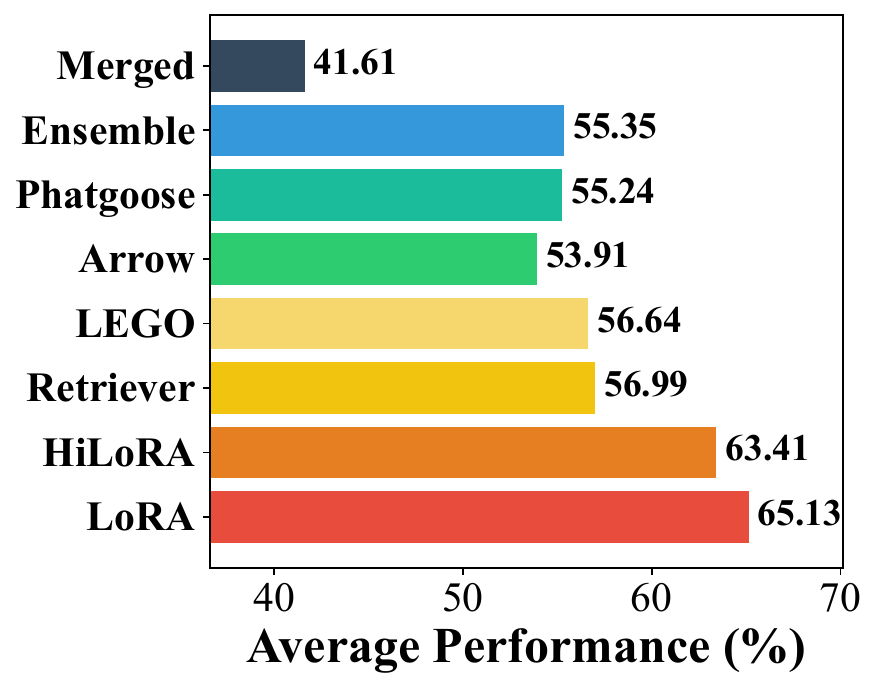}
    \vspace{-9mm}
    \caption{Average accuracy over ten NLI tasks, with five seen tasks and five unseen tasks. \texttt{HiLoRA} achieves the best performance and approaches the accuracy of task-specific LoRAs. Detailed results are shown in Tab.~\ref{tab:within_cluster}.}
    \label{fig:acc}
    \vspace{-3mm}
\end{wrapfigure}

Recent work has attempted to address the above challenges by integrating Mixture-of-Experts (MoE) mechanisms with LoRAs~\citep{ge2025dynamic}, where gating functions are designed to route inputs to a subset of LoRAs.
However, these gating functions often rely on explicit task labels~\citep{ma2024modula} or require gradient-based training of additional gating parameters~\citep{muqeeth2024learning}, which restricts their applicability in practical deployment. 
Moreover, most methods rely on top-$k$ gating scores~\citep{ostapenko2024towards,zhao2024loraretriever}, which lead to either excessive or insufficient activations and thus limit adaptability.
In parallel, some studies focus on LoRA merging, which integrates multiple task-specific LoRAs into a single unified module to enhance cross-domain generalization by leveraging knowledge across tasks~\citep{coleman2024adaptive,zhao2025merging}.
These approaches impose a uniform architecture across tasks, which limits flexibility and degrades performance in scenarios involving diverse tasks.
A more detailed discussion of related work is provided in Appendix~\ref{sec:related}.
This motivates the following research question:

\textbf{\textit{Can we adaptively leverage a large collection of specialized LoRA modules to support both seen and unseen tasks without retraining or explicit task labels?}}

In this paper, we highlight \textbf{three key observations} about the structure of LoRA, derived from empirical analysis and experimental evidence. (i) Each rank-one direction in a LoRA is formed by pairing a row vector from the down-projection matrix with a corresponding column vector from the up-projection matrix. Since these directions function independently, one can treat each pair as a \textit{rank-one component (ROC)}, which serves as the basic unit of LoRA.
(ii) Within a LoRA, the down-projection vectors across ROCs exhibit strong randomness and primarily serve as scaling factors that modulate the effect of the corresponding up-projection vectors.
(iii) In contrast, the up-projection vectors show clear clustering patterns, often forming multiple groups within the same LoRA. These clusters capture distinct semantic aspects of the LoRA’s adaptive capacity.

Building on these insights, we propose \texttt{HiLoRA}, a hierarchical LoRA routing framework designed to adaptively support robust domain generalization. To the best of our knowledge, \texttt{HiLoRA} is the first method to introduce hierarchical routing at the granularity of ROCs, while also providing theoretical guarantees for LoRA identification through error bounds. At the sequence level, \texttt{HiLoRA} narrows the candidate space and improves robustness by activating only a subset of LoRAs based on input-LoRA similarity. To enable comparison between inputs and LoRAs that reside in different parameter spaces, each LoRA is represented as a Gaussian distribution fitted to a small set of sampled embeddings, and similarity is measured using Gaussian likelihoods. This probabilistic formulation not only allows reliable distinction between seen and unseen tasks, but also provides confidence signals that guide the adaptive determination of both the number of activated LoRAs and their ROC allocation. At the token level, the down-projection vectors within ROCs are used to further select the most informative ROCs, refining routing without introducing additional parameters or requiring training. 
We summarize our contributions as follows.
\begin{itemize}[noitemsep, topsep=0pt, left=.3em]
\item \textbf{New Insight.} We identify the ROC as the fundamental semantic unit of LoRA and show both the feasibility and necessity of performing routing at this fine-grained granularity.
\item \textbf{Hierarchical LoRA Routing Framework.} \texttt{HiLoRA} constructs a dynamic LoRA pool, where each LoRA is represented as a Gaussian distribution fitted from samples of its training dataset. At the sequence level, the Gaussian likelihood scores between the input and LoRAs are calculated. The maximum score determines both the number of activated LoRAs and the overall ROC budget, while normalized scores guide probabilistic sampling for ROC allocation. At the token level, routing is further refined by selecting ROCs with stronger down-projection responses.
\item \textbf{Theoretical Guarantee.} We derive error bounds for LoRA identification, providing the first formal guarantees that \texttt{HiLoRA} preserves the corresponding LoRAs for seen tasks and the closest LoRAs for unseen tasks with high probability, thereby ensuring robust routing across domains.
\item \textbf{Experimental Performance.} As shown in Fig.~\ref{fig:acc} for a representative case, \texttt{HiLoRA} consistently outperforms state-of-the-art baselines in both within-cluster and cross-cluster evaluations, achieving accuracy gains of up to {\small $55\%$} on LLaMA2-7B and {\small $13\%$} on FLAN-T5-large, while maintaining practical inference throughput.
\end{itemize}

\section{Preliminaries}\label{sec_preli}
    \textbf{{Basic Formulation of LoRA}.}
    LoRA~\citep{hu2022lora} achieves performance comparable to full fine-tuning by freezing the pretrained weights {\small $\boldsymbol{W}_0$} and inserting trainable low-rank matrices {\small $\Delta \boldsymbol{W}$} into selected layers, yielding {\small $\boldsymbol{W}’ = \boldsymbol{W}_0 + \Delta \boldsymbol{W}$}.
    The update matrix is factorized as {\small $\Delta \boldsymbol{W} = \boldsymbol{B}\boldsymbol{A}$}, where {\small $\boldsymbol{A} \in \mathbb{R}^{r \times d}$} is the down-projection matrix and {\small $\boldsymbol{B} \in \mathbb{R}^{d \times r}$} is the up-projection matrix, with rank {\small $r \ll d$}.
    This reduces the number of trainable parameters from {\small $d^2$} to {\small $rd$} while retaining strong adaptability.
    Given an input {\small $\boldsymbol{x} \in \mathbb{R}^d$}, the sub-module output {\small $\boldsymbol{y} \in \mathbb{R}^d$}, originally computed as {\small $\boldsymbol{y} = \boldsymbol{W}_0 \boldsymbol{x}$}, is reformulated under LoRA adaptation as:
    
    {\small
    \vspace{-5mm}
    \begin{equation}
       \boldsymbol{y} = \boldsymbol{W}_0 \boldsymbol{x} + \Delta \boldsymbol{W} \boldsymbol{x} = \boldsymbol{W}_0 \boldsymbol{x} + \boldsymbol{B}\boldsymbol{A}\boldsymbol{x}.
    \end{equation}
    \vspace{-5mm}}

    \textbf{{Dyadic Product Representation}.} 
    Let {\small $\{\boldsymbol{a}_i^\top\}_{i=1}^r$} denote the set of row vectors of {\small $\boldsymbol{A}$}
    and {\small $\{\boldsymbol{b}_i\}_{i=1}^r$} denote the set of column vectors of {\small $\boldsymbol{B}$}, where {\small $\boldsymbol{a}_i, \boldsymbol{b}_i \in \mathbb{R}^{d}$}.
     Under this notation, the low-rank update can be written as
    {\small $\Delta \boldsymbol{W} = \boldsymbol{B}\boldsymbol{A} = \sum_{i=1}^r \big(\boldsymbol{b}_i \boldsymbol{a}_i^\top\big)$}, which expresses {\small $\Delta \boldsymbol{W}$} as a sum of {\small $r$} dyadic products, each formed by the outer product of two vectors {\small $(\boldsymbol{a}_i, \boldsymbol{b}_i)$}.
    Substituting this representation into the forward computation yields:

    {\small
    \vspace{-5mm}
    \begin{equation}
        \boldsymbol{y} = \boldsymbol{W}_0 \boldsymbol{x} + {\textstyle\sum_{i=1}^r} \big(\boldsymbol{b}_i \boldsymbol{a}_i^\top\big)\boldsymbol{x}.
    \end{equation}
    \vspace{-5mm}}

    In this decomposition, each row of the down-projection matrix {\small $\boldsymbol{A}$} is paired with the corresponding column of the up-projection matrix {\small $\boldsymbol{B}$}. The pair {\small $(\boldsymbol{a}_i, \boldsymbol{b}_i)$} acts as an indivisible unit, which we define as a rank-one component (ROC). A ROC corresponds to one rank in LoRA and serves as the fundamental element of its adaptive capacity. 
    Consequently, the ROC constitutes the minimal routing unit of LoRA, and we next introduce an adaptive strategy to determine both the number and the selection of ROCs to activate for each input.

\section{Methodology}\label{sec_metho}

\subsection{HiLoRA Framework}
    \textbf{{Problem Formulation}.} Consider a pre-trained LLM {\small $\boldsymbol{L}$} and a pool of {\small $I$} task-specific LoRAs, denoted as {\small $\Phi = \{\phi_1, \phi_2, \dots, \phi_I\}$}. 
    It is implemented by inserting low-rank matrices into selected layers of {\small $\boldsymbol{L}$}.
    For clarity, the low-rank parameters of {\small $\phi_i$} at a given layer are denoted as {\small $\boldsymbol{A}_i$} and {\small $\boldsymbol{B}_i$}, with rank {\small $r_i$}.
    Our objective is to design a \textit{routing mechanism} that exploits the pool of LoRAs {\small $\Phi$} without requiring additional training or explicit task labels.
    Such a mechanism should perform competitively on tasks with corresponding LoRAs available in the pool (\textit{seen tasks}), while also generalizing to inputs from domains lacking specialized LoRAs (\textit{unseen tasks}).

    \begin{figure}[!htb]
    	\vspace{-4mm}
    	\centering
    	\includegraphics[width=1\textwidth]{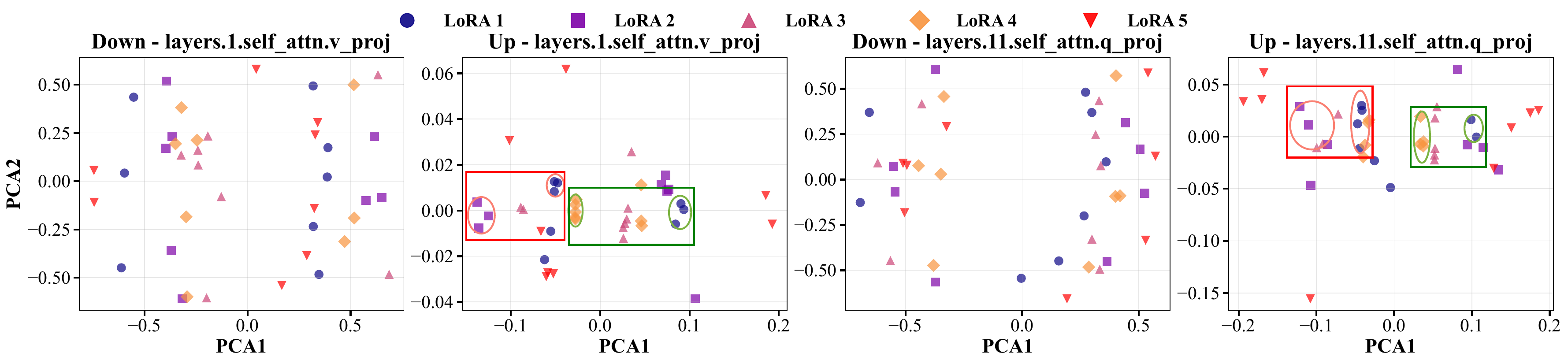}
    	\vspace{-9mm}
    	\caption{Scatter plots of the first two principal components derived from vectors in LoRA projection matrices specialized for five NLI tasks. The boxes highlight examples where optimal routing for an unseen task (pink) would involve selecting only the vectors aligned with relevant semantics.}
    	\label{fig:motivation}
    	\vspace{-3mm}
    \end{figure}

%\subsection{Motivating Observations}
    \label{subsection:observation}

    \textbf{{Motivating Observations}.} Empirical findings in~\citep{zhu2024asymmetry} indicate that the down-projection matrix {\small $\boldsymbol{A}$} primarily extracts features from the input, while the up-projection matrix {\small $\boldsymbol{B}$} transforms these features to generate the output.  
    Accordingly, the down-projection vector {\small $\boldsymbol{a}$}  in each ROC determines how strongly the input aligns with its direction, and this value regulates the effect of the paired up-projection vector {\small $\boldsymbol{b}$}.
    To further validate this distinction and examine additional properties of ROCs, we visualize LoRA parameters using Principal Component Analysis (PCA)~\citep{abdi2010principal}. In particular, vectors obtained by slicing the projection matrices along the rank dimension, {\em i.e.}, {\small $\{\boldsymbol{a}_i, \boldsymbol{b}_i\}_{i=1}^r$}, are projected into a two-dimensional space.  We analyze five LoRAs fine-tuned on different NLI tasks, with the resulting scatter plots shown in Fig.~\ref{fig:motivation}, where vectors sharing the same color and shape are drawn from the same LoRA.
    To ensure that the reported observations are not limited to these cases, additional visualizations are provided in the Appendix~\ref{sub_app_observation}.

    Three key observations arise from these visualizations.
    (i) The down-projection vectors of ROCs exhibit strong randomness and show little alignment with task semantics. This confirms that down-projection vector {\small $\boldsymbol{a}$} primarily functions as a scaling factor, rather than encoding domain-specific information.
    (ii) In contrast, the up-projection vectors of ROCs within a given LoRA exhibit clear task-dependent patterns. These vectors often form multiple distinct clusters, with each cluster representing a different semantic fragment of the LoRA’s adaptive capacity.
    (iii)  For domain generalization, activating an entire LoRA introduces parameter redundancy and interference, since unrelated clusters are involved simultaneously. 
    Taken together, these observations suggest that effective routing should selectively activate only those clusters or vectors aligned with relevant semantics. 
    As illustrated in Fig.~\ref{fig:motivation}, when the pink LoRA corresponds to an unseen task, the optimal routing selectively activates only specific clusters ({\em e.g.}, the red box selects purple and blue clusters, while the green box selects orange and blue clusters). Similarly, in the fourth subfigure, the activated ROCs originate from the purple, blue, and orange clusters, although the precise cluster assignments differ.

\begin{wrapfigure}{r}{0.5\linewidth}
    \vspace{-3mm}
    \centering
    \includegraphics[width=\linewidth]{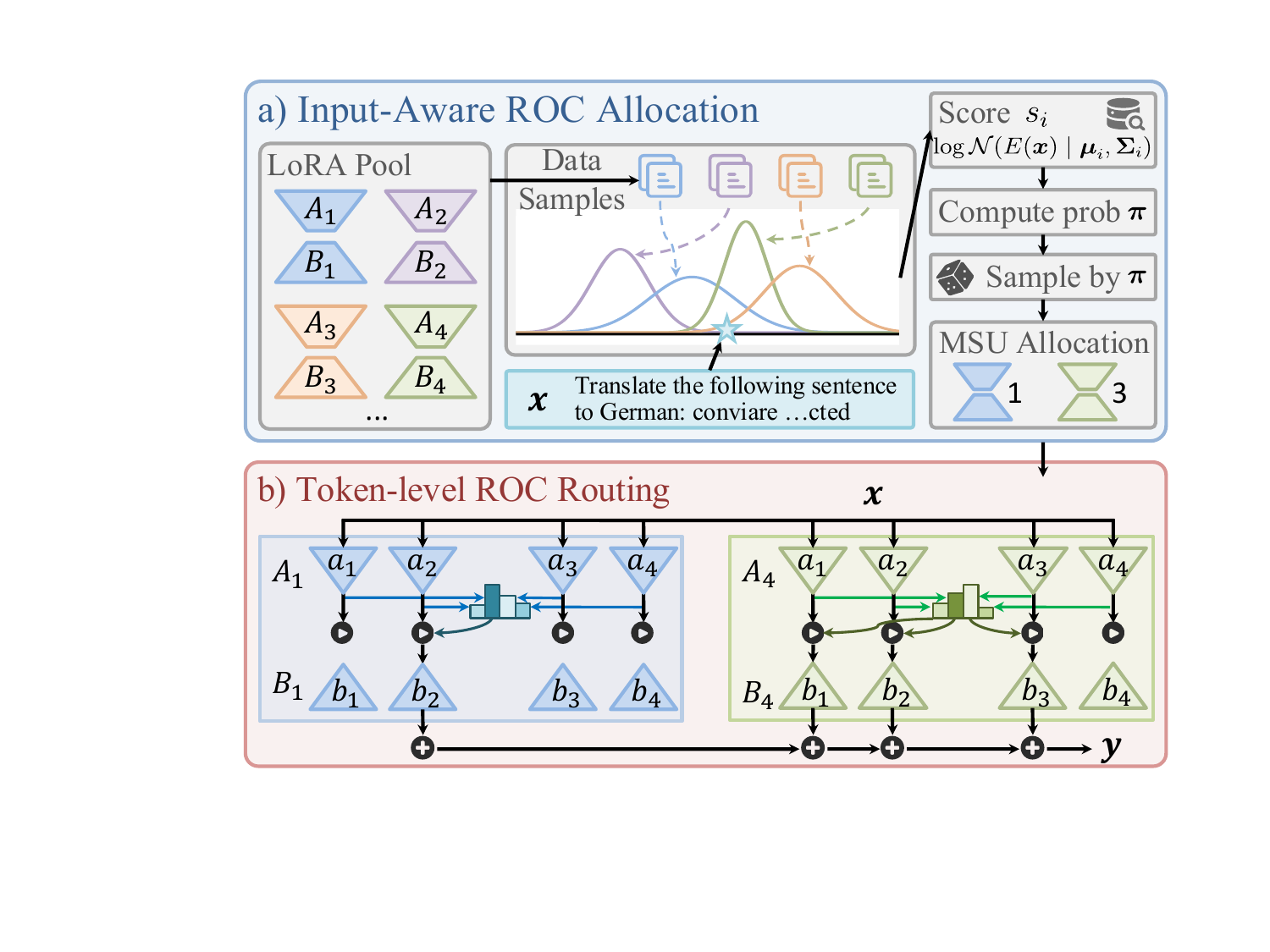}
    \vspace{-6mm}
    \caption{Overview of \texttt{HiLoRA} architecture.}
    \label{fig:model}
    \vspace{-3mm}
\end{wrapfigure}

\textbf{{Workflow of \texttt{HiLoRA}}.} 
Motivated by these observations, routing at the granularity of ROCs is highly desirable. 
However, directly selecting ROCs from the entire LoRA pool faces two main challenges.
First, the candidate space is excessively large, which makes exhaustive selection computationally infeasible.
Second, the space is noisy, as ROCs from different LoRAs vary in relevance and quality, making it difficult to evaluate them under a unified criterion.
To address these issues, we introduce \texttt{HiLoRA}, an adaptive hierarchical routing framework over a pool of task-specific LoRAs designed to achieve training-free domain generalization.
Given an input sequence {$\boldsymbol{x}$}, \texttt{HiLoRA} operates in two stages.
(i) \emph{Input-Aware ROC Allocation}: At the sequence level, the framework measures the similarity between {\small $\boldsymbol{x}$} and each LoRA {\small $\phi_i$} using Gaussian likelihoods.
Based on these probabilistic similarities, it selects a subset of LoRAs and assigns an appropriate number of ROCs to each.
(ii) \emph{Token-Level ROC Routing}: At the token level, the framework further refines adaptation by dynamically routing each token in {\small $\boldsymbol{x}$} to the most relevant ROCs within the subset of LoRAs selected in stage (i). 
In both stages, comparisons are performed under a unified criterion, which ensures fair evaluation across LoRAs and their ROCs.
The overview of our framework \texttt{HiLoRA} is illustrated in Fig.~\ref{fig:model}.

\subsection{Input-Aware ROC Allocation}
At the sequence level, the goal is to identify candidate LoRAs from the pool and 
allocate a suitable number of ROCs to each, according to their relevance to the input. 
A key challenge arises because the input representations and LoRA parameters reside in distinct spaces, which prevents direct comparison.
To address this issue, inspired by retrieval-based methods, each LoRA can be represented by a small set of samples drawn from its training dataset \citep{zhao2024loraretriever}. 
Instead of embedding LoRAs and inputs into a shared space and computing cosine similarity, we approximate each LoRA with a Gaussian distribution fitted to the sampled embeddings. This yields a probabilistic representation that enables more robust matching \citep{cha2021advance,li2023modeling}.
This probabilistic representation provides an information-theoretic characterization: inputs from seen tasks attain high likelihood under their corresponding LoRA distributions, while inputs from unseen tasks can still be aligned by evaluating their likelihood across all source distributions. Moreover, the resulting probabilities guide stochastic allocation of ROCs, which not only reduces over-reliance on a single LoRA but also encourages exploration across multiple relevant candidates.

Formally, let {\small $E$} denote a sentence embedding model and {\small $\boldsymbol{c}$} denote an instruction. 
The instructed embedding of an input {\small $\boldsymbol{x}$} is given by
{\small $\boldsymbol{z} = E(\boldsymbol{c} \oplus \boldsymbol{x})$},
where {\small $\oplus$} denotes concatenation.
Following \citet{zhao2024loraretriever}, we set the instruction to
“Represent the sentence for similar task retrieval’’ to encourage sequence-level similarity.
For each LoRA module {\small $\phi_i$}, we randomly sample $m$ domain-specific examples, 
obtain their instructed embeddings 
{\small $\{\boldsymbol{z}^{(i)}_1, \dots, \boldsymbol{z}^{(i)}_m\}$}, and fit a Gaussian distribution:

{\small 
\vspace{-3mm}
    \begin{equation}
    p_i(\boldsymbol{z}) = \mathcal{N}(\boldsymbol{z} \mid \boldsymbol{\mu}_i, \boldsymbol{\Sigma}_i), 
    \text{ where }\boldsymbol{\mu}_i = \frac{1}{m} {\textstyle \sum_{j=1}^m} \boldsymbol{z}^{(i)}_j, \text{ }
    \boldsymbol{\Sigma}_i = \frac{1}{m-1}{\textstyle \sum_{j=1}^m} \big(\boldsymbol{z}^{(i)}_j - \boldsymbol{\mu}_i\big)\big(\boldsymbol{z}^{(i)}_j - \boldsymbol{\mu}_i\big)^\top.
\end{equation}
\vspace{-5mm}}

For a given input {\small $\boldsymbol{x}$}, we then compute its log-likelihood under each LoRA distribution as the similarity score:
{\small $s_i(\boldsymbol{x}) = (1/\tilde{d})\log p_i(\boldsymbol{z}), \forall i \in \{1,\dots, I\}$},
where {\small $\tilde{d}$} denotes the embedding dimension.
Since inputs may come from either seen or unseen tasks, with seen tasks typically producing higher scores. Therefore, two cases are considered depending on whether a positive score is present:
{\small \begin{equation}
\label{eq:lora}
    \mathbb{C}(\boldsymbol{x})= 
            \begin{cases}
                \{i \mid s_i(\boldsymbol{x})>0\}, & \text{if } \max_i s_i(\boldsymbol{x})>0, \\
                \arg \text{top}_i^c\ s_i(\boldsymbol{x}), & \text{if } \max_i s_i(\boldsymbol{x})\le 0,
            \end{cases}
\end{equation}}

\vspace{-2mm}
where {\small $c = \max\{|\max_i s_i(\boldsymbol{x})|, k\}$}.       
If the maximum score is positive, only LoRAs with positive scores are retained. Otherwise, the Top-$c$ LoRAs are selected, ensuring that at least {\small $k$} candidates remain while allowing more to be included when the maximum score is strongly negative. This design reflects the intuition that lower maximum scores signal weak alignment with all LoRAs, and thus broader coverage improves robustness in domain alignment.
The total ROC budget is defined as
{\small $O(\boldsymbol{x}) = \gamma \cdot \sum_{i \in \mathbb{C}(\boldsymbol{x})} r_i,$}
where {\small $\gamma \in (0,1)$} is a scaling factor. A large value of {\small $\gamma$} may introduce redundancy and interference, whereas a small value may exclude essential information.
Thus, {\small $\gamma$} is set to balance accuracy and efficiency by activating a compact yet sufficient set of ROCs.
The scores of selected LoRAs are then normalized into probabilities: {\small $\pi_i(\boldsymbol{x}) =  \frac{\exp(s_i(\boldsymbol{x}))}{\sum_{j \in \mathbb{C}(\boldsymbol{x})} \exp(s_j(\boldsymbol{x}))}, \forall i \in \mathbb{C}(\boldsymbol{x})$}.
Using these probabilities, the ROC allocation {\small $\{o_i\}_{i \in \mathbb{C}(\boldsymbol{x})}$} is sampled from a multinomial distribution with parameters {\small $O(\boldsymbol{x})$} and {\small $\boldsymbol{\pi}(\boldsymbol{x})$}, subject to the per-LoRA capacity constraint {\small $o_i \le r_i$}.

\subsection{Token-level ROC Routing}
\textbf{{ROC Routing within Chosen LoRAs}.}
At the token level, routing is refined by operating on the granularity of ROCs.
As discussed in Sec.~\ref{subsection:observation}, the down-projection vectors mainly act as scaling factors.
Therefore, the projection value {\small $\boldsymbol{a}^\top\boldsymbol{x}$} provides a natural criterion for ROC selection, with larger values indicating stronger relevance between the token and the corresponding ROC. 
This criterion helps reduce redundancy by prioritizing the most informative ROCs while filtering out those with limited contribution or potential interference.
Formally, for each layer and each token, and for every LoRA {\small $i \in \mathbb{C}(\boldsymbol{x})$} selected at the sequence level, we compute the projection values {\small $\boldsymbol{A}_i \boldsymbol{x}$}. The most informative ROCs are then identified by selecting the indices of the top-$o_i$ components ranked by projection value:
{\small $\mathbb{J}_i=\arg \text{top}_j^{o_i}(\boldsymbol{a}_{ij}^\top \boldsymbol{x})$}.
The LoRA output for this layer is then obtained by aggregating the contributions of all activated ROCs: {\small $ \boldsymbol{y}' = \sum_{i \in \mathbb{C}(\boldsymbol{x})} \sum_{j \in \mathbb{J}_i} \boldsymbol{b}_{ij} \, (\boldsymbol{a}_{ij}^\top \boldsymbol{x})$}.

It is important to emphasize that this routing introduces no additional parameters or retraining. Since projection values {\small $\boldsymbol{a}^\top\boldsymbol{x}$} are required for all activated ROCs, the only extra computation arises from evaluating projections of ROCs that are ultimately not selected. This overhead is minimal compared to the overall forward pass, ensuring efficiency while preserving robust adaptation.

\textbf{{Variance Normalization for Adaptive ROCs}.} 
In \texttt{HiLoRA}, the number of activated ROCs is adaptive and may range from {\small $1$} to {\small $\sum_{i=1}^I r_i$}, where {\small $r_i$} is the rank of LoRA {\small $\phi_i$}.
This variability can cause fluctuations in the scale of the aggregated LoRA output, which in turn may reduce the stability of model performance. 
Empirical findings in \citep{zhao2025merging} show that LoRA outputs are approximately distributed as zero-mean Gaussians, with variance that grows with the number of activated ROCs.
To mitigate this effect, we normalize the aggregated output by a scaling factor
{\small $\sqrt{\overline{r}(\boldsymbol{x})/O(\boldsymbol{x})}$}, 
where {\small $\overline{r}(\boldsymbol{x}) = \frac{1}{|\mathbb{C}(\boldsymbol{x})|}\sum_{i \in \mathbb{C}(\boldsymbol{x})} r_i$} is the average rank of the selected LoRAs~\citep{vaswani2017attention}.
Therefore, the output of a given layer for input {\small $\boldsymbol{x}$} becomes: 
{\small $\boldsymbol{y} = \boldsymbol{W}_0\boldsymbol{x} + \sqrt{\overline{r}/{O(\boldsymbol{x})}} \sum_{i \in \mathbb{C}(\boldsymbol{x})}\sum_{j \in \mathbb{J}_i}\boldsymbol{b}_{ij}^\top(\boldsymbol{a}_{ij}^\top)\boldsymbol{x}$}.
This variance normalization property has been formally established in Theorem 3.1 of \citep{zhao2025merging}. For clarity and completeness, we restate it as a Lemma~\ref{lem:variance} in Appendix~\ref{sec_app}.

\subsection{Theoretical Analysis}
We present the error bounds of LoRA identification in \texttt{HiLoRA} under two scenarios: (i) \textit{in-distribution (ID)} inputs from seen tasks, and (ii) \textit{out-of-distribution (OOD)} inputs from unseen tasks.

\textbf{{Error Bound for ID Inputs}.} 
For inputs from seen tasks, we provide a Top-$k$ error bound that measures the probability of the corresponding LoRA being excluded from the selected set.

\begin{Lemma}
\label{lemma_binary}%\citep{nielsen2014generalized}.} 
For any two distributions {\small $i,j$} with class-conditional Gaussians {\small $\mathcal{N}(\boldsymbol{\mu}_i,\boldsymbol{\Sigma}_i)$} and {\small $\mathcal{N}(\boldsymbol{\mu}_j,\boldsymbol{\Sigma}_j)$} and prior probabilities {\small $\pi_i,\pi_j$}, the Bayes error rate satisfies:  
{\small $P_{\rm err}^{(2)}(i,j) \;\le\; \sqrt{\pi_i\pi_j}\; \exp\!\big(-B_{ij}\big)$},  
where  
{\small $B_{ij} = \tfrac18(\boldsymbol{\mu}_i-\boldsymbol{\mu}_j)^\top \Big(\tfrac{\boldsymbol{\Sigma}_i + \boldsymbol{\Sigma}_j}{2}\Big)^{-1} (\boldsymbol{\mu}_i-\boldsymbol{\mu}_j) \;+\; \tfrac12 \log \frac{\big|(\boldsymbol{\Sigma}_i + \boldsymbol{\Sigma}_j)/{2}\big|}{\sqrt{|\boldsymbol{\Sigma}_i||\boldsymbol{\Sigma}_j|}}$}.
\end{Lemma}

\vspace{-2mm}
In this paper, priors are not incorporated in the score. The same derivation yields the simplified form  
{\small $P_{\rm err}^{(2)}(i,j)\;\le\;\exp(-B_{ij})$}. Based on this Lemma, we have the following error bound.  

\begin{Theorem}\label{thm:topk}
For an input {\small $\boldsymbol{x}$} with true label {\small $t_i$}, the prediction is determined by the top-$k$ scores  
{\small $s_i(\boldsymbol{x})$}.  
The probability that the LoRA corresponding to {\small $t_i$} is not included in the Top-$k$ set {\small $\mathbb{K}$} is bounded as:  
{\small \begin{equation}
    \Pr\big(i \notin \mathbb{K}\big)\ 
    \;\le\; \tfrac{1}{k}{\textstyle\sum_{j\ne i}} \exp\!\big(-B_{ij}\big).
\end{equation}}
\end{Theorem}

\vspace{-2mm}
Theorem~\ref{thm:topk} shows that for ID inputs, the probability of excluding the correct LoRA decreases in two ways:
(1) it drops exponentially as task distributions become more separable (larger {\small $B_{ij}$}); and
(2) it decreases proportionally with the size of the Top-$k$ set.

\textbf{{Error Bound for OOD Inputs}.}
For an input {\small $\boldsymbol{x}$} from unseen tasks, no exact task-specific LoRA exists in the pool, suppose it comes from an unknown target distribution {\small $q$}. 
Define the information-theoretically closest source domain as {\small $i^\star := \arg\min_{i \in \{1,\cdots,I\}}\, D_{\mathrm{KL}}(q\,\|\,p_i)$}.

\begin{Theorem}\label{thm:ood}

Let the prediction be based on the top-$k$ scores 
{\small $s_i(\boldsymbol{x})$}. 
For any {\small $\alpha\in(0,1]$} and
{\small $M_\alpha^j = \boldsymbol{\Sigma}_q^{-1} + \alpha \boldsymbol{\Sigma}_j^{-1} - \alpha \boldsymbol{\Sigma}_{i^\star}^{-1} \ \succ\ 0$}, the probability that the LoRA {\small $i^\star$} is excluded from the Top-$k$ set {\small $\mathbb{K}$} satisfies:
{\small
\vspace{-2mm}
\begin{equation}
    \Pr\!\big(i^\star \notin \mathbb{K}\big)\ 
    \;\le\; \tfrac{1}{k}\,{\textstyle\sum_{j\ne i^\star}} \;
    C_\alpha^j \, |M_\alpha^j|^{-1/2}\,
    \exp\!\left(
        \tfrac{1}{2}\,(h_\alpha^j)^\top (M_\alpha^j)^{-1} h_\alpha^j \;-\; K_\alpha^j
    \right),
\end{equation}
\vspace{-5mm}}

where
{\small $h_\alpha^j = \boldsymbol{\Sigma}_q^{-1}\boldsymbol{\mu}_q + \alpha \boldsymbol{\Sigma}_j^{-1}\boldsymbol{\mu}_j - \alpha \boldsymbol{\Sigma}_{i^\star}^{-1}\boldsymbol{\mu}_{i^\star}, \quad
K_\alpha^j = \tfrac{1}{2}\boldsymbol{\mu}_q^\top \boldsymbol{\Sigma}_q^{-1}\boldsymbol{\mu}_q
+ \tfrac{\alpha}{2}(
    \boldsymbol{\mu}_j^\top \boldsymbol{\Sigma}_j^{-1}\boldsymbol{\mu}_j
    - \boldsymbol{\mu}_{i^\star}^\top \boldsymbol{\Sigma}_{i^\star}^{-1}\boldsymbol{\mu}_{i^\star}), \quad
C_\alpha^j = \exp(
    -\tfrac{\alpha}{2}\log|\boldsymbol{\Sigma}_j|
    + \tfrac{\alpha}{2}\log|\boldsymbol{\Sigma}_{i^\star}|
    - \tfrac{1}{2}\log|\boldsymbol{\Sigma}_q|
).$}
\end{Theorem}

%\vspace{-2mm}
Here, {\small $M_\alpha^j$} is a weighted precision matrix combining the covariance information of $q$, $j$, and $i^\star$, while the condition {\small $M_\alpha^j \succ 0$} guarantees that the quadratic form is well-defined and divergence is finite; {\small $h_\alpha^j$} is a mean–precision vector measuring the displacement of $q$ relative to $j$ and $i^\star$ under covariance-adjusted weighting; {\small $K_\alpha^j$} is a correction term involving second-order statistics, capturing quadratic differences in alignment; {\small $C_\alpha^j$} is a scale factor derived from covariance determinants, quantifying relative volume mismatch.
Theorem~\ref{thm:ood} shows that for OOD inputs, the probability of excluding the closest LoRA decreases in two ways:
(1) it drops exponentially when the unseen distribution {\small $q$} is better aligned with {\small $i^\star$} and more distinct from other source domains {\small $j$};
(2) it decreases proportionally with the size of the Top-$k$ set.

\textbf{{Remarks}.}
Theorem~\ref{thm:topk} and Theorem~\ref{thm:ood} highlight two key insights.
(i) When domains are well separated and the LoRA pool spans diverse tasks, the error bounds are tight, ensuring strong guarantees in both ID and OOD cases. This condition is often met in practice, as task domains are generally distinguishable, and open-source repositories already provide a rich collection of LoRAs across diverse tasks.
(ii) Increasing $k$ tightens the bound, but excessively large values introduce redundancy and interference. To balance this trade-off, \texttt{HiLoRA} adaptively adjusts the size of the activated set based on input-LoRA similarity, retaining the corresponding or closest LoRA with high probability while avoiding unnecessary overhead and parameter interference.

\section{Experiments}\label{sec_exp}
%In this section, we present a series of experiments and provide a concise interpretation of the results to evaluate HiLoRA.
\subsection{Experimental Setup}%Implementation Details}
\textbf{{Datasets and Models}.}
We use a subset of tasks from FLAN-v2~\citep{wei2022finetuned}, and organize them into ten clusters: Natural Language Inference (NLI), Question Answering (QA), Sentiment Analysis, Translation, Commonsense Reasoning, Paraphrase, Struct-to-Text, Coreference Resolution, Text Correction, and Word-level tasks, following the categorization in \citet{wei2022finetuned}.
We construct the LoRA pool by downloading task-specific LoRAs for the selected tasks from HuggingFace. 
Since evaluation metrics vary across tasks, we adopt task-dependent measures including accuracy, F1 score, BLEU, and ROUGE-{1, 2, L}. Details of the selected tasks, their grouping, and metrics are provided in Appendix~\ref{subsec_dataset}. As backbone models, we use two representative LLMs: LLaMA2-7B~\citep{touvron2023llama} and FLAN-T5-large~\citep{chung2024scaling}.

\textbf{{Baselines}.}
We compare \texttt{HiLoRA} with the following state-of-the-art methods.
(i) \texttt{HiLoRA-GS}: a variant of \texttt{HiLoRA} that applies only sequence-level routing.
(ii) \texttt{HiLoRA-ROC}: a variant of \texttt{HiLoRA} that applies only token-level routing by ranking all ROCs across LoRAs and selecting the top-$k$.
(iii) \texttt{Retriever} \citep{zhao2024loraretriever}: a sequence-level method that retrieves the top-$k$ LoRAs based on cosine similarity between input and LoRA embedding.
(iv) \texttt{LEGO} \citep{zhao2025merging}: a ROC-level merging method that clusters all ROCs into $k$ groups, merges each cluster into a new ROC, and applies the merged clusters to all tasks.
(v) \texttt{Arrow} \citep{ostapenko2024towards}: a token-level routing approach that builds gating vectors from the first right singular vector of the LoRA update {\small $\boldsymbol{BA}$}.
(vi) \texttt{Phatgoose} \citep{muqeeth2024learning}: a token-level routing method where gating vectors are trained separately for each task.
(vii) \texttt{Ensemble} \citep{muhlematter2024lora}: an ensemble method that combines all LoRAs by averaging their outputs.
(viii) \texttt{Merged} \citep{ostapenko2023case}: a method where all LoRAs are merged into a single module shared across tasks.

\textbf{{Implementation Details}.}
We set the inference batch size to {\small $32$}. For each seen task, {\small $m=20$} domain-specific samples from the corresponding dataset are used to fit a Gaussian distribution. The sentence embedding model {\small $E$} is implemented with instructor-base~\citep{su2023one}, an instruction-tuned encoder that produces task-aware representations. The scaling factor {\small $\gamma$} is fixed at {\small $40\%$}. 
Following \citep{zhao2024loraretriever}, we set the parameter {\small $k=3$} for all LoRA-level routing methods, and correspondingly {\small $k=24$} for all ROC-level routing methods. All experiments are conducted in PyTorch on a system with Ubuntu 22.04, Intel Xeon Platinum 8558P processors (192 CPUs), 2.0~TiB of memory, and NVIDIA H100 GPUs with {\small 80GB} memory.

\subsection{Main Results}
Experimental results are reported under two evaluation settings: (i) the within-cluster setting evaluates performance when test tasks originate from the same cluster as the seen tasks, and (ii) the cross-cluster setting measures generalization to tasks from unseen clusters.

\vspace{-5mm}
\begin{table}[htbp]
\centering
\caption{Performance on the NLI cluster using 
LLaMA2-7B and FLAN-T5-large. Tasks with a white background are set as \textit{seen} tasks, while those with a gray background are set as \textit{unseen} tasks.
For each task, the best accuracy among all methods is in \textbf{bold}, and the second best is \underline{underlined}.}
\label{tab:within_cluster}
\begin{adjustbox}{max width=\textwidth}
\begin{tabular}{l | c | c c c c c c c c c}
\toprule
\textbf{Methods} & \textbf{LoRA} & \textbf{HiLoRA} & \textbf{HiLoRA-GS} & \textbf{HiLoRA-ROC} & \textbf{Retriever} & \textbf{LEGO} & \textbf{Arrow} & \textbf{Phatgoose} & \textbf{Ensemble} & \textbf{Merged} \\
\midrule
\multicolumn{11}{c}{\textit{LLaMA2-7B}}\\
\midrule
ANLI-r1    & 46.40 & \textbf{45.00} & \underline{42.10} & 38.90 & 36.10 & 37.00 & 38.90 & 37.00 & 35.80 & 31.70  \\
\rowcolor{gray!20}
\cellcolor{white}ANLI-r2    & \cellcolor{white}40.10 & \textbf{40.60} & \underline{38.70} & 36.20 & 36.40 & 37.70 & 36.40 & 36.40 & 36.80 & 32.60  \\
ANLI-r3    & 36.92 & \textbf{37.67} & 36.17 & 35.92 & 35.25 & 34.75 & \underline{36.25} & 35.42 & 34.50 & 31.08 \\
\rowcolor{gray!20}
\cellcolor{white}CB         & \cellcolor{white}80.00 &68.00 & \underline{70.00} & 64.00 & 66.00 & 66.00 & 64.00 & \textbf{74.00} & \underline{68.00} & 56.00 \\
MNLI       & 77.66 & \textbf{76.33} & 74.06 & 70.78 & \underline{74.22} & 71.91 & 60.51 & 62.58 & 67.66 & 39.92 \\
\rowcolor{gray!20}
\cellcolor{white}MNLI-mis   & \cellcolor{white}79.69 & \textbf{78.59} & 74.69 & 69.38 & \textbf{75.78} & 71.80 & 60.82 & 62.34 & 68.75 & 40.59 \\
QNLI       & 77.27 & \textbf{78.28} & \underline{77.23} & 59.02 & 62.19 & 58.71 & 59.02 & 59.80 & 57.89 & 45.23  \\
\rowcolor{gray!20}
\cellcolor{white}RTE        & \cellcolor{white}72.96 & 74.44 & \textbf{75.56} & 65.93 & 65.93 & 71.11 & \underline{75.56} & 74.07 & 71.48 & 53.70  \\
SNLI       & 67.42 & \underline{69.45} & 68.13 & 69.34 & \textbf{70.94} & 67.46 & 59.06 & 57.89 & 62.58 & 35.27 \\
\rowcolor{gray!20}
\cellcolor{white}WNLI       & \cellcolor{white}72.86 & \textbf{65.71} & \underline{62.29} & 48.57 & 47.14 & 50.00 & 48.57 & 52.86 & 50.00 & 50.00 \\
\midrule
Avg        & 65.13 & \textbf{63.41} & \underline{61.89} & 55.80 & 56.99 & 56.64 & 53.91 & 55.24 & 55.35 & 41.61  \\
\midrule
\multicolumn{11}{c}{\textit{FLAN-T5-Large}}\\
\midrule
Avg        & 67.81 & \textbf{67.70} & 64.85 & 66.53 & \underline{66.76} & 56.20 & 57.81 & 55.29 & 56.19 & 53.03 \\
\bottomrule
\end{tabular}
\end{adjustbox}
\end{table}
\vspace{-2mm}

\textbf{{Within-cluster Setting}.}
In this setting, experiments are conducted on ten NLI tasks, with half designated as seen tasks and the other half as unseen tasks.  
Results are summarized in Tab.~\ref{tab:within_cluster}, while per-task accuracy for T5 is provided in the Appendix~\ref{subsec:result} due to page limits.
From the table, it can be observed that the proposed \texttt{HiLoRA} substantially outperforms all baselines on both \textit{seen} and \textit{unseen} tasks, improving average accuracy by {\small $6$}-{\small $22\%$} on LLaMA and up to {\small $14\%$} on T5-large.
More specifically: (i) On \textit{seen} tasks, \texttt{HiLoRA} achieves performance comparable to the oracle setting (LoRA in Tab.~\ref{tab:within_cluster}) where each input is served by its task-specific LoRA, and in some cases even surpasses it, {\em e.g.} ANLI-r3 and QNLI. This indicates that \texttt{HiLoRA} not only identifies the task-specific LoRA corresponding to the given input but also leverages useful ROCs from other LoRAs to further enhance performance.
(ii) On \textit{unseen} tasks, \texttt{HiLoRA} also delivers consistently strong results, demonstrating its ability to generalize by aligning inputs with semantically related LoRAs and refining predictions through selective ROC activation.
(iii) The gains are particularly notable on LLaMA, which relies more heavily on LoRA adaptation than T5-large. Since T5-large has already been extensively pretrained on FLAN-style tasks, the relative contribution of LoRA adaptation is smaller compared to LLaMA.
Methods such as \texttt{Retriever}, \texttt{Arrow}, \texttt{Phatgoose}, and \texttt{Ensemble} activate a fixed number of LoRAs (or even all of them) without accounting for conflicts or redundancies among ROCs, leading to parameter interference or insufficiency that ultimately degrades performance.
\texttt{LEGO}, while incorporating ROC clustering and merging, remains input-agnostic and retains all clusters, thereby failing to eliminate parameter redundancy.
The \texttt{Merged} baseline performs worst due to severe parameter interference when all LoRAs are combined into a single module.
In contrast, \texttt{HiLoRA} employs a hierarchical routing strategy: at the sequence level, it prunes irrelevant LoRAs via Gaussian similarity sampling, and at the token level, it selects only the most effective ROCs. This design reduces parameter redundancy and prevents interference,  and explains the consistent performance gains observed across both seen and unseen tasks.

\vspace{-5mm}
\begin{table}[htbp]
\centering
\caption{Performance of LLaMA2-7B and FLAN-T5-large under the cross-cluster setting. 
For tasks with multiple evaluation metrics, the average score across metrics is computed first, and the cluster score is then obtained by averaging over all tasks in the cluster. 
For each cluster, the best result among all methods is in \textbf{bold}, and the second best is \underline{underlined}.}
\label{tab:cross_cluster}
\begin{adjustbox}{max width=\textwidth}
\centering
\begin{tabular}{l | c| c c c c c c c c c}
\toprule
\textbf{Methods} & \textbf{LoRA} & \textbf{HiLoRA} & \textbf{HiLoRA-GS} & \textbf{HiLoRA-ROC} & \textbf{Retriever} & \textbf{LEGO} & \textbf{Arrow} & \textbf{Phatgoose} & \textbf{Ensemble} & \textbf{Merged} \\
\midrule
\multicolumn{11}{c}{\textit{LLaMA2-7B}}\\
\midrule
NLI        & 63.13 & \textbf{46.54} & 44.23 & \underline{45.00} & 43.78 & 42.89 & 42.29 & 43.78 & 43.57 & 11.69 \\
QA         & 59.66 & \textbf{46.95} & 43.56 & 43.19 & 43.55 & \underline{46.67} & 39.37 & 45.10 & 44.89 & 10.09 \\
Senti.  & 59.87 & \textbf{54.43} & 49.88 & 54.00 & 50.12 & 52.93 & 40.76 & \underline{53.00} & 50.26 &  4.19 \\
Trans.& 21.98 & 20.78 & \textbf{21.80} & 14.92 &  9.50 & 16.45 & \underline{20.93} & 20.47 & 20.77 & 11.91 \\
Common.& 67.11 & \textbf{52.76} & 50.27 & 51.29 & 44.99 & 50.14 & 50.83 & 50.88 & \underline{52.03} & 15.24 \\
Paraph. & 66.88 & \underline{53.08} & 50.11 & 42.73 & \textbf{54.51} & 39.91 & 45.09 & 47.31 & 49.06 &  7.61 \\
StT & 44.51 & \textbf{28.31} & \underline{28.18} & 24.86 & 27.32 & 15.89 & 27.71 & 28.01 & 27.21 & 24.94 \\
Corefe.& 47.95 & \underline{61.59} & \textbf{62.04} & 59.30 & 59.02 & 58.79 & 61.04 & 58.23 & 60.70 &  6.98 \\
Text-Corr. & 54.73 & \underline{30.98} & \textbf{33.21} & 25.73 & 26.14 & 24.04 & 29.35 & 29.58 & 29.93 &  6.34 \\
Word       & 67.02 & \underline{46.13} & 45.51 & 43.08 & \textbf{46.73} & 38.61 & 45.73 & 45.43 & 43.09 & 11.47 \\
\midrule
\multicolumn{11}{c}{\textit{FLAN-T5-Large}}\\
\midrule
NLI        & 67.81 & \textbf{63.49} & 58.65 & \underline{63.21} & 62.04 & 52.18 & 50.59 & 62.08 & 52.75 & 49.11 \\
QA         & 67.39 & \textbf{63.44} & 61.73 & 63.08 & 60.87 & 60.03 & 59.40 & \underline{63.13} & 60.39 & 58.51 \\
Senti.  & 59.18 & \textbf{58.55} & 58.14 & \underline{58.49} & 57.73 & 58.11 & 57.96 & 58.13 & 58.00 & 57.94 \\
Trans.& 18.97 & 18.79 & \underline{18.80} & 18.55 & \textbf{18.88} & 18.77 & 18.74 & 18.61 & 18.77 & 18.65 \\
Paraph.& 78.33 & \textbf{75.18} & 74.91 & 68.00 & 72.52 & 73.63 & 72.85 & \underline{74.76} & 73.97 & 72.96 \\
StT & 60.18 & \underline{59.85} & 59.83 & 59.42 & \textbf{59.88} & 59.80 & 59.79 & 59.76 & 59.79 & 59.75 \\
Corefe.& 63.13 & \textbf{63.89} & 61.63 & \underline{63.61} & 62.04 & 60.95 & 60.95 & 62.07 & 62.04 & 60.68 \\
Text-Corr. & 54.91 & \textbf{54.83} & 54.21 & 53.68 & 54.01 & 54.56 & 54.45 & \underline{54.68} & 54.63 & 54.21 \\
Word       & 71.55 &73.35 & 72.22 & 64.01 & 72.10 & \underline{73.86} & 72.59 & \textbf{73.63} & 73.91 & 73.40 \\
\bottomrule
\end{tabular}
\end{adjustbox}
\end{table}
\vspace{-3mm}

\textbf{{Cross-cluster Setting}.}  
In this setting, each cluster is treated as unseen in turn, while the remaining clusters serve as seen. For LLaMA2-7B, the LoRA pool contains all 50 task-specific modules, while for T5-large, only 33 modules are included due to the limited availability of community-provided LoRAs. 
Performance is evaluated on all tasks within the unseen cluster, with average results reported in Tab.~\ref{tab:cross_cluster} and detailed metrics provided in Appendix~\ref{subsec:result}.
This configuration is more challenging than the within-cluster settings, as unseen tasks may differ substantially in semantics from the seen ones. Nevertheless, \texttt{HiLoRA} achieves strong cross-domain generalization, yielding accuracy gains of up to {\small $55\%$} on LLaMA2-7B and {\small $13\%$} on T5-large.   Although it does not always attain the highest score in every cluster, its performance is consistently within {\small $2.5\%$} of the best and remains superior to all baselines.
These results highlight the routing capability of \texttt{HiLoRA}, which mitigates parameter redundancy and interference even when adapting to previously unseen clusters. 
Interestingly, \texttt{Ensemble} performs relatively better in this setting than in the within-cluster case, since activating a larger number of LoRAs helps capture broader information, which is beneficial for serving tasks from unseen clusters. 
These observations further highlight the advantage of \texttt{HiLoRA}, which adaptively determines the number of activated LoRAs according to input-LoRA similarity, thereby preserving sufficient information while avoiding redundancy as formalized in Eq. (\ref{eq:lora}).

\subsection{Further Analysis}

\begin{wrapfigure}{r}{0.45\linewidth}
    \centering
    \vspace{-5mm}
    \includegraphics[width=\linewidth]{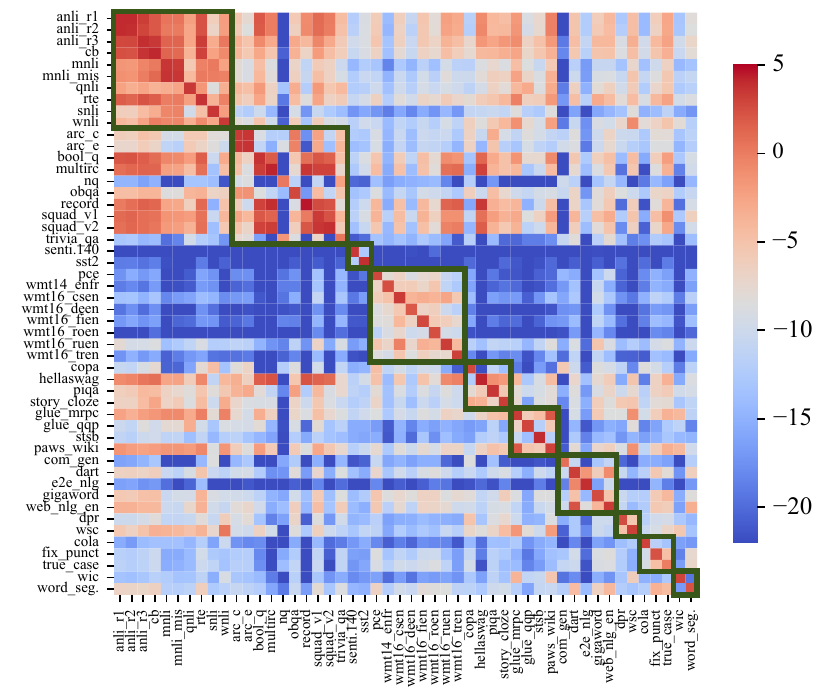}
    \vspace{-8mm}
    \caption{Input-LoRA similarity heatmap produced by \texttt{HiLoRA}, where tasks from the same cluster are enclosed within green boxes for clarity.}
    \label{fig:mapping}
    \vspace{-3mm}
\end{wrapfigure}

\textbf{Performance of Input Mapping.}  
To evaluate the input routing capability of \texttt{HiLoRA}, we visualize the similarities among task embeddings across different tasks. Fig.~\ref{fig:mapping} presents a heatmap, where tasks from the same cluster are grouped by \textit{green boxes}.  
Three observations can be made: 
(i) Task embeddings within the same domain exhibit higher similarity, indicating that \texttt{HiLoRA} effectively captures relationships across related tasks.  
(ii) The similarity values exhibit a substantially broader range ({\small $-22$} to {\small $5$}) compared with the narrower interval of {\small $-1$} to {\small $1$} obtained by \texttt{Retriever} \citep{zhao2024loraretriever} (see Appendix~\ref{subsec:result}).  This broader contrast sharpens intra-cluster cohesion while maintaining clear separation across clusters, thereby improving task alignment and reducing the risk of mismatching semantically different tasks.
(iii) Unlike other methods, \texttt{HiLoRA} adaptively determines the number of activated LoRAs based on input-LoRA similarity, ({\em i.e.}, Eq. (\ref{eq:lora})). 
As shown in Fig.~\ref{fig:mapping}, for easy cases such as seen tasks, \texttt{HiLoRA} activates only {\small $1$}-{\small $2$} LoRAs, whereas in the cross-cluster setting it scales up to {\small $11$} LoRAs to handle more dissimilar tasks.
This dynamic adaptation and flexibility reduces redundancy while ensuring sufficient coverage.
Consequently, to sustain robust performance, the LoRA pool requires a number of modules and reasonable task coverage.

\begin{wrapfigure}{r}{0.55\textwidth}  
    \centering
    \vspace{-3mm}
    \begin{tabular}{cc}
		\hspace{-5mm}
		\begin{minipage}[t]{0.48\linewidth}
			\centering            \includegraphics[width=1.0\textwidth]{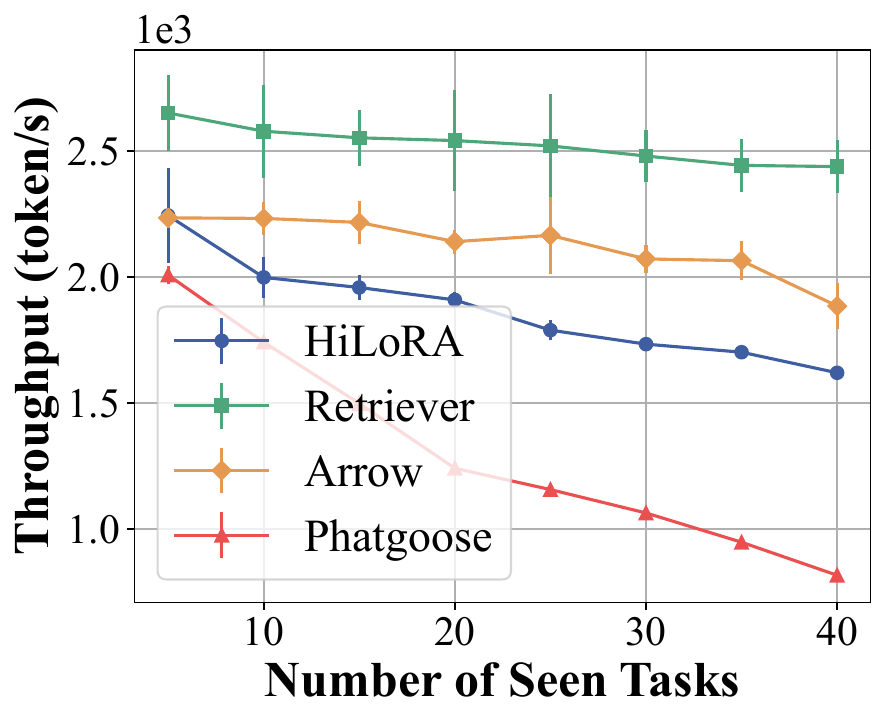}	
			\vspace{-7mm}
			\caption{Inference throughput with different numbers of seen tasks. }
			\label{fig:throughput}
		\end{minipage}
		\hspace{1mm}
		\begin{minipage}[t]{0.48\linewidth}
			\centering
			\includegraphics[width=1.0\textwidth]{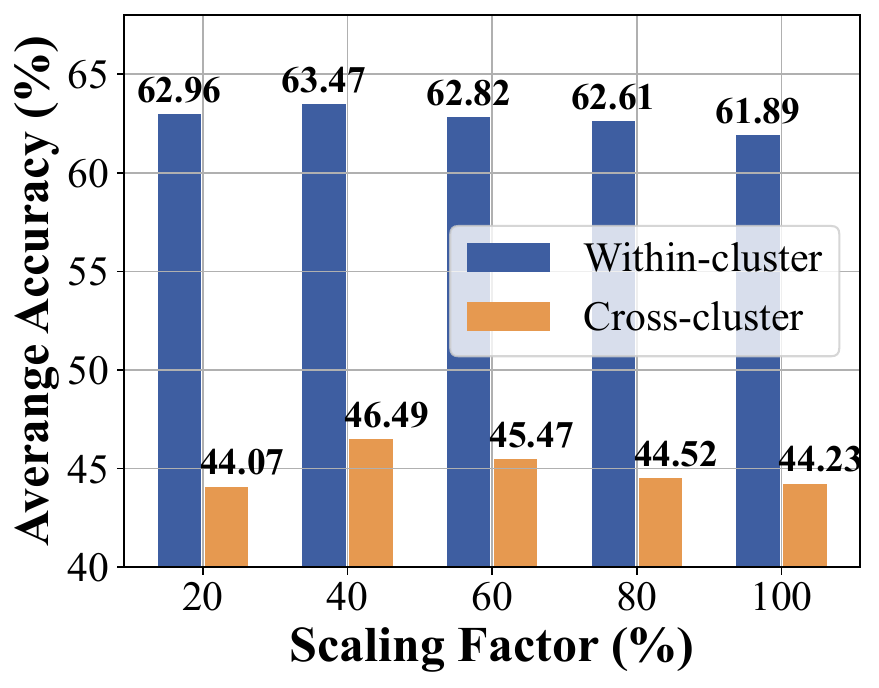}	
			\vspace{-7mm}
			\caption{Performance of \texttt{HiLoRA} with different value of scaling factor {\small $\gamma$}. }
			\label{fig:ablation}
		\end{minipage}
	\end{tabular}
    \vspace{-3mm}
\end{wrapfigure}

\textbf{{Inference Throughput}.}  
We further evaluate inference throughput by comparing \texttt{HiLoRA} with dynamic routing baselines under different numbers of seen tasks, ranging from {\small $5$} to {\small $40$}. For each configuration, inference throughput is measured on a test set containing {\small $5$} seen tasks and {\small $5$} unseen tasks. As shown in Fig.~\ref{fig:throughput}, throughput decreases gradually as the number of seen tasks increases. 
Compared with some single-level routing methods, \texttt{HiLoRA} incurs a throughput reduction of about {\small $7$}-{\small $30\%$}, but still achieves up to {\small $90\%$} higher throughput compared to \texttt{Phatgoose}.
Considering the substantial performance gains observed in both within-cluster and cross-cluster settings, this moderate reduction in throughput is acceptable and highlights a favorable balance between efficiency and accuracy.

\textbf{{Ablation Study}.}
Here, we conduct an ablation study on the scaling factor {\small $\gamma$}, which controls the number of total ROCs activated. Experiments are performed under both within-cluster and cross-cluster settings, and the evaluation covers all NLI tasks. As shown in Fig.~\ref{fig:ablation}, setting {\small $\gamma=40\%$} yields the best overall performance. 
Larger values of {\small $\gamma$} activate excessive ROCs, introducing parameter redundancy and interference that ultimately reduce performance. In particular, setting {\small $\gamma=100\%$} corresponds to \texttt{HiLoRA-GS}, which relies solely on global sequence-level routing. Conversely, smaller values may exclude too many informative ROCs, leading to insufficient representation capacity and degraded performance.
These results suggest that maintaining a moderate number of ROCs is crucial. Moreover, this balance not only improves prediction accuracy but also helps control inference cost, as fewer ROCs need to be processed. Overall, the ablation results confirm that an appropriate scaling factor is key to ensuring both efficiency and robust performance in \texttt{HiLoRA}.
\section{Conclusion}\label{sec-conclu}
In this paper, we present \texttt{HiLoRA}, a training-free framework for adaptive hierarchical routing over pools of task-specific LoRAs to support robust domain generalization. \texttt{HiLoRA} builds on structural insights into LoRA by treating each ROC as the minimal routing unit. At the sequence level, it adaptively selects candidate LoRAs and allocates ROCs using Gaussian likelihoods, narrowing the search space and improving robustness. At the token level, routing is further refined by selecting the most informative ROCs, which reduces redundancy and alleviates interference.
Theoretical analysis and extensive experiments demonstrate that \texttt{HiLoRA} reliably identifies relevant LoRAs, substantially improves domain generalization, and maintains efficiency with only a moderate reduction in inference throughput.

Despite its strengths, \texttt{HiLoRA} has several limitations. It relies on a small number of task-specific samples to construct Gaussian representations, which may not always be accessible and could raise privacy concerns. Moreover, the token-level routing mechanism is empirically validated but lacks formal theoretical guarantees. Future research could focus on addressing these limitations to broaden the practical applicability of the approach.

\newpage
\section*{Reproducibility Statement}
We have made extensive efforts to ensure the reproducibility of our work. The formulation of \texttt{HiLoRA}, including the definition of rank-one components and the hierarchical routing framework, is described in detail in Sec.~\ref{sec_metho}, with complete theoretical analyses and proofs provided in Appendix~\ref{sec_app}. All datasets and task clusters are drawn from widely used public benchmarks, and the corresponding preprocessing steps and evaluation protocols are fully documented in Appendix~\ref{subsec_dataset}. Experimental configurations, including hyperparameter choices, routing parameters, and hardware settings, are reported in Sec.~\ref{sec_exp}, and additional empirical results are provided in Appendix~\ref{subsec:result}. To further support reproducibility and enable reuse, we will release source code and scripts for dataset preparation upon publication.

\bibliography{iclr2026_conference}
\bibliographystyle{iclr2026_conference}

\newpage
\appendix
\section{Related Work}\label{sec:related}
Recent advances in extending LoRA for cross-domain adaptation fall into two primary directions: MoE-style routing and LoRA merging.

\textbf{MoE-style Routing.} These methods extend LoRA adaptation by dynamically activating subsets of LoRAs through gating functions~\citep{mao2025survey}. At the sequence level, routing is performed using task-level similarity or global gating scores to select LoRA experts for the entire input, as in MoA~\citep{feng2024mixture} and MoLE~\citep{wu2024mixture}. At the token level, methods such as LoRA-Switch~\citep{kong2024lora} and Arrow~\citep{ostapenko2024towards} introduce token-wise gating to activate different LoRAs for different positions. Hybrid strategies combine these two levels, {\em e.g.}, HMoRA~\citep{liao2025hmora} and MoLoRA~\citep{hou2025molora}, aiming to balance efficiency and flexibility. 
Beyond entire LoRA routing,  rank-level routing has also been explored, where each rank is treated as a micro-expert and subsets are activated, as in SMoRA~\citep{zhao2025each}.
Although these methods demonstrate the benefits of dynamic expert selection, they exhibit two key limitations: (i) they typically require training additional gating parameters, which undermines scalability and hinders deployment in training-free scenarios, and (ii) they impose a fixed activation budget, which reduces adaptability when handling diverse or unseen tasks.
In contrast, our work introduces a hierarchical routing framework that performs training-free selection at the sequence level and further refines routing at the ROC level, enabling finer-grained control that reduces redundancy and improves robustness across both seen and unseen domains.

\textbf{{LoRA Merging}.}
These methods aim to combine multiple task-specific LoRAs into a single unified module to support domain generalization~\citep{huang2024lorahub,coleman2024adaptive,qorbani2025semantic}.
ZipLoRA achieves effective style and subject composition by directly merging independently trained LoRAs~\citep{shah2024ziplora} for vision and text generation.
LoRA-LEGO introduces rank-wise clustering and re-assembly of LoRA ranks to construct merged adapters with adjustable capacity~\citep{zhao2025merging}.
Beyond heuristic merging, recent works explore more principled strategies: Closed-Form Merging (LoRM) derives analytical solutions for merging parameter-efficient modules in federated continual learning settings~\citep{salami2025closedform}, while Adaptive LoRA Merge with Parameter Pruning further enhances robustness in low-resource domains by combining merging with pruning and lightweight fine-tuning~\citep{miyano2025adaptive}.
While these approaches enhance cross-domain generalization by leveraging knowledge across tasks, they enforce a one-size-fits-all merged model. This limits flexibility and often degrades performance in scenarios involving diverse or unseen tasks.
Our work addresses these limitations by designing an adaptive routing framework that adaptively selects LoRAs at the sequence level and refines the choice at the ROC level, providing task-aware composition while reducing redundancy and interference.
\section{Theoretical Demonstration}\label{sec_app}
\subsection{ariance Normalization Property}

For completeness, we restate the variance normalization property, originally established as Theorem~3.1 in \citet{zhao2025merging}. As the full proof is already provided in the cited work, we omit the derivation here and present the result in the form of a lemma below.

\begin{Lemma} [Theorem 3.1 in \citep{zhao2025merging}]
\label{lem:variance}
    Let {\small $\boldsymbol{A}_{1} \in \mathbb{R}^{d \times r}$}, {\small $\boldsymbol{B}_{1} \in \mathbb{R}^{r \times d}$},  and {\small $\boldsymbol{A}_{2} \in \mathbb{R}^{d \times k}$}, {\small $\boldsymbol{B}_{2} \in \mathbb{R}^{k \times d}$}, where all entries are independently sampled from the standard normal distribution {\small $\mathcal{N}(0,1)$}. If the product {\small $\boldsymbol{A}_{2}\boldsymbol{B}_{2}$} is rescaled by the factor {\small $\sqrt{r/k}$}, then the variance of the entries in {\small $\boldsymbol{A}_{1}\boldsymbol{B}_{1}$} coincides with that of the normalized product: 
    {\small $\mathrm{Var}(\boldsymbol{A}_{1}\boldsymbol{B}_{1}) \;=\; \mathrm{Var}\!\left(\sqrt{\tfrac{r}{k}}\,\boldsymbol{A}_{2}\boldsymbol{B}_{2}\right)$}.
\end{Lemma}

\subsection{Proof of Lemma~\ref{lemma_binary}}

The Bayes error for the optimal (MAP) decision rule is:
\begin{align*}
P_{\rm err}^{(2)}(i,j)
&=\int \min\{\pi_i p_i(\boldsymbol{z}),\;\pi_j p_j(\boldsymbol{z})\}\,d\boldsymbol{z}
\;\le\;\int \sqrt{\pi_i p_i(\boldsymbol{z})\;\pi_j p_j(\boldsymbol{z})}\,d\boldsymbol{z}\\
&=\sqrt{\pi_i\pi_j}\int \sqrt{p_i(\boldsymbol{z})\,p_j(\boldsymbol{z})}\,d\boldsymbol{z}
\;=\;\sqrt{\pi_i\pi_j}\,\rho(p_i,p_j),
\end{align*}
where the first inequality follows from the elementary bound {\small $\min{a,b} \leq \sqrt{ab}$} for {\small $a,b \geq 0$}, and {\small $\rho(p_i,p_j) := \int \sqrt{p_i p_j}$} denotes the Bhattacharyya coefficient (affinity) between $p_i$ and $p_j$.

For {\small $k\in\{i,j\}$}, the Gaussian densities are:
\[
p_k(\boldsymbol{z})=(2\pi)^{-d/2}|\boldsymbol{\Sigma}_k|^{-1/2}
\exp\!\Big(-\tfrac12(\boldsymbol{z}-\boldsymbol{\mu}_k)^\top \boldsymbol{\Sigma}_k^{-1}(\boldsymbol{z}-\boldsymbol{\mu}_k)\Big).
\]
Thus, we have:
\[
\sqrt{p_i(\boldsymbol{z})p_j(\boldsymbol{z})}=(2\pi)^{-d/2}|\boldsymbol{\Sigma}_i|^{-1/4}|\boldsymbol{\Sigma}_j|^{-1/4}
\exp\!\Big(-\tfrac12(\boldsymbol{z}-\tilde{\boldsymbol{\mu}})^\top\tilde{\boldsymbol{\Sigma}}^{-1}(\boldsymbol{z}-\tilde{\boldsymbol{\mu}})\Big)\,e^{-C},
\]
where {\small $\tilde{\boldsymbol{\Sigma}}^{-1}=\tfrac12(\boldsymbol{\Sigma}_i^{-1}+\boldsymbol{\Sigma}_j^{-1}), \tilde{\boldsymbol{\mu}}=\tilde{\boldsymbol{\Sigma}}\cdot\tfrac12(\boldsymbol{\Sigma}_i^{-1}\boldsymbol{\mu}_i+\boldsymbol{\Sigma}_j^{-1}\boldsymbol{\mu}_j)$}, with
{\small $C=\tfrac14\!\left(\boldsymbol{\mu}_i^\top\boldsymbol{\Sigma}_i^{-1}\boldsymbol{\mu}_i+\boldsymbol{\mu}_j^\top\boldsymbol{\Sigma}_j^{-1}\boldsymbol{\mu}_j-2\tilde{\boldsymbol{\mu}}^\top\tilde{\boldsymbol{\Sigma}}^{-1}\tilde{\boldsymbol{\mu}}\right)$}.

Integration yields:
\[
\rho(p_i,p_j)=|\boldsymbol{\Sigma}_i|^{-1/4}|\boldsymbol{\Sigma}_j|^{-1/4}\,e^{-C}\,|\tilde{\boldsymbol{\Sigma}}|^{1/2}.
\]

Using standard matrix identities, we have:
\[
C=\tfrac18(\boldsymbol{\mu}_i-\boldsymbol{\mu}_j)^\top\Big(\tfrac{\boldsymbol{\Sigma}_i+\boldsymbol{\Sigma}_j}{2}\Big)^{-1}(\boldsymbol{\mu}_i-\boldsymbol{\mu}_j)
,\quad
|\tilde{\boldsymbol{\Sigma}}|^{1/2}|\boldsymbol{\Sigma}_i|^{-1/4}|\boldsymbol{\Sigma}_j|^{-1/4}
 = exp(-\tfrac12\log\frac{\big|\tfrac{\boldsymbol{\Sigma}_i+\boldsymbol{\Sigma}_j}{2}\big|}{\sqrt{|\boldsymbol{\Sigma}_i||\boldsymbol{\Sigma}_j|}}).
\]

Substituting them into {\small $\rho(p_i,p_j)$} gives:
\[
\rho(p_i,p_j)=
\exp\!\left(
-\tfrac18(\boldsymbol{\mu}_i-\boldsymbol{\mu}_j)^\top\!\Big(\tfrac{\boldsymbol{\Sigma}_i+\boldsymbol{\Sigma}_j}{2}\Big)^{-1}(\boldsymbol{\mu}_i-\boldsymbol{\mu}_j)
-\tfrac12\log\frac{\big|\tfrac{\boldsymbol{\Sigma}_i+\boldsymbol{\Sigma}_j}{2}\big|}{\sqrt{|\boldsymbol{\Sigma}_i||\boldsymbol{\Sigma}_j|}}
\right)
=\exp(-B_{ij}).
\]

Therefore, we proved:
\[
P_{\rm err}^{(2)}(i,j)\;\le\;\sqrt{\pi_i\pi_j}\;\rho(p_i,p_j)
=\sqrt{\pi_i\pi_j}\,\exp\!\big(-B_{ij}\big).
\]

%\paragraph{Remarks.}
%(i) The above is the Bhattacharyya bound; optimizing the exponent over a mixing parameter $\alpha\in[0,1]$ yields the tighter Chernoff bound.
%(ii) In the special case $\boldsymbol{\Sigma}_i=\boldsymbol{\Sigma}_j=\boldsymbol{\Sigma}$, the determinant term cancels and $B_{ij}=\tfrac18\|\boldsymbol{\mu}_i-\boldsymbol{\mu}_j\|^2_{\boldsymbol{\Sigma}^{-1}}$.

\subsection{Proof of Theorem~\ref{thm:topk}}
\textbf{\textit{Pairwise Error}.} For a given input {\small $\boldsymbol{x}$} and {\small $label(x) = t_i$}, define the pairwise overtake events as {\small $A_{ij}=\{p_j(\boldsymbol{z})\ge p_i(\boldsymbol{z})\}, \quad j\neq i$}. 
For {\small $A_{ij}$}, we have:
\begin{align}\label{eq_aij}
    \Pr(A_{ij}\mid label(x)=t_i) &\;=\; \int_{\{p_j(\boldsymbol{z}) \ge p_i(\boldsymbol{z})\}} p_i(\boldsymbol{z})\,d\boldsymbol{z} 
 \;=\; \int_{\{p_j(\boldsymbol{z}) \ge p_i(\boldsymbol{z})\}} \min\{p_i(\boldsymbol{z}),p_j(\boldsymbol{z})\}d\boldsymbol{z} \\ \nonumber
 %&\;\le\;\int \min\{p_i,p_j\}d\boldsymbol{z} = \rho(p_i,p_j) = exp(-B_{ij}), \\ \nonumber
 &\;\le\; \int \min\{p_i,p_j\}\,d\boldsymbol{z} \le\; \int \sqrt{p_ip_j}\; dx =  \rho(p_i,p_j) = \exp(-B_{ij}),
\end{align}

where the first equality follows from the definition of the error event: under class $i$, misclassification occurs precisely when $p_j(\boldsymbol{z})\geq p_i(\boldsymbol{z})$;
the second equality holds because, on the region $\{p_j(\boldsymbol{z})\geq p_i(\boldsymbol{z})\}$, we have $\min\{p_i(\boldsymbol{z}),p_j(\boldsymbol{z})\}=p_i(\boldsymbol{z})$;
the inequality is obtained by extending the domain of integration;
and the last two equality uses the Bhattacharyya coefficient, as established in Lemma~\ref{lemma_binary}.

\textbf{\textit{Top-$k$ Error}.} Let {\small $N_1\ =\ \sum_{j\ne i}\mathbf 1_{A_{ij}}.$} denote the number of rivals that beat {\small $i$}.
Then the Top-$k$ error event under {\small $label(\boldsymbol{x}) = i$} is {\small $\{N_1\ge k\}$}. Now, we have the following analysis:

\begin{align*}
    \Pr(N_1\ge k\mid label(x)=t_i) &\ \le\ \frac{\mathbb E[N_1\mid label(x)=t_i]}{k}    \\
    &\ = \ \frac{1}{k}\sum_{j\ne i}\Pr(A_{ij}\mid label(x)=t_i) \\
&\ \le\ \frac{1}{k}\sum_{j\ne i}exp(-B_{i,j}),
\end{align*}
where the first inequality applies the Markov’s inequality; the equality follows from computing $\mathbb{E}[N_1]$ and substituting the pairwise error terms; and the final inequality then uses the bound in Eq.~(\ref{eq_aij}).

% general bound for non-negative random variables, $\Pr(X \ge t) \le \mathbb{E}[X]/t$ for $t>0$; 
\subsection{Proof of Theorem~\ref{thm:ood}}
For any competitor {\small $j\neq i^\star$}, consider the event 
{\small $ p_j(\boldsymbol{z})\ge  p_{i^\star}(\boldsymbol{z})$}. 
For any {\small $\alpha\in(0,1]$}, the Markov-Chernoff technique gives:
\[
\Pr_{\boldsymbol{z}\sim q}\!\big( p_j(\boldsymbol{z})\ge  p_{i^\star}(\boldsymbol{z})\big)
=\Pr\!\Bigg(\Big(\tfrac{p_j(\boldsymbol{z})}{p_{i^\star}(\boldsymbol{z})}\Big)^{\!\alpha}\ge 1\Bigg)
\;\le\; \mathbb{E}_{q}\!\left[\Big(\tfrac{p_j(\boldsymbol{z})}{p_{i^\star}(\boldsymbol{z})}\Big)^{\!\alpha}\right],
\]

where The first equality holds because, for any {\small $\alpha>0$}, the event 
{\small $\{p_j \ge p_{i^\star}\}$} is equivalent to {\small $\Big\{\Big(\tfrac{p_j}{p_{i^\star}}\Big)^{\!\alpha} \ge 1 \Big\}$}; and the first inequality then follows from Markov’s inequality: if a random variable 
{\small $X \ge 0$}, then for any {\small $t>0$},
{\small $\Pr(X \ge t) \;\le\; \frac{\mathbb{E}[X]}{t}$}.

\begin{Lemma}
    \label{lem:qweighted}
Let $q=\mathcal{N}(\boldsymbol{\mu}_q,\boldsymbol{\Sigma}_q)$, $p_j=\mathcal{N}(\boldsymbol{\mu}_j,\boldsymbol{\Sigma}_j)$, and $p_{i^\star}=\mathcal{N}(\boldsymbol{\mu}_{i^\star},\boldsymbol{\Sigma}_{i^\star})$ be full-rank $d$-variate Gaussians.
For any $\alpha\in(0,1]$, assume
{\small $M_\alpha^j\;:=\;\boldsymbol{\Sigma}_q^{-1}+\alpha\,\boldsymbol{\Sigma}_j^{-1}-\alpha\,\boldsymbol{\Sigma}_{i^\star}^{-1}\ \succ\ 0$}.

Then the $\alpha$-moment of the likelihood ratio admits the closed form as follows:
\[
\mathbb E_{\boldsymbol{z}\sim q}\!\left[\Big(\tfrac{p_j(\boldsymbol{z})}{p_{i^\star}(\boldsymbol{z})}\Big)^{\!\alpha}\right]
\;=\; C_\alpha^j\;\,|M_\alpha^j|^{-1/2}\,
\exp\!\left(\tfrac12\,(h_\alpha^j)^\top (M_\alpha^j)^{-1} h_\alpha^j \,-\, K_\alpha^j\right),
\]
where {\small $
h_\alpha^j=\boldsymbol{\Sigma}_q^{-1}\boldsymbol{\mu}_q+\alpha\,\boldsymbol{\Sigma}_j^{-1}\boldsymbol{\mu}_j-\alpha\,\boldsymbol{\Sigma}_{i^\star}^{-1}\boldsymbol{\mu}_{i^\star},
K_\alpha^j=\tfrac12\boldsymbol{\mu}_q^\top\boldsymbol{\Sigma}_q^{-1}\boldsymbol{\mu}_q+\tfrac{\alpha}{2}\big(\boldsymbol{\mu}_j^\top\boldsymbol{\Sigma}_j^{-1}\boldsymbol{\mu}_j-\boldsymbol{\mu}_{i^\star}^\top\boldsymbol{\Sigma}_{i^\star}^{-1}\boldsymbol{\mu}_{i^\star}\big),
C_\alpha^j=\exp\!\Big(-\tfrac{\alpha}{2}\log|\boldsymbol{\Sigma}_j|+\tfrac{\alpha}{2}\log|\boldsymbol{\Sigma}_{i^\star}|-\tfrac{1}{2}\log|\boldsymbol{\Sigma}_q|\Big)$}.
\end{Lemma}

\textbf{Proof.}
By Chernoff/Markov's trick (see \citep{chernoff1952measure}), we have:
\[
\mathbb E_q\!\left[\Big(\tfrac{p_j(\boldsymbol{z})}{p_{i^\star}(\boldsymbol{z})}\Big)^{\!\alpha}\right]
=\int_{\mathbb R^d} q(\boldsymbol{z})\exp\!\Big(\alpha\big(\log p_j(\boldsymbol{z})-\log p_{i^\star}(\boldsymbol{z})\big)\Big)\,d\boldsymbol{z}.
\]
Write each log-density in quadratic form:
\(
\log p(\boldsymbol{z})=-\tfrac d2\log(2\pi)-\tfrac12\log|\boldsymbol{\Sigma}|-\tfrac12(\boldsymbol{z}-\boldsymbol{\mu})^\top\boldsymbol{\Sigma}^{-1}(\boldsymbol{z}-\boldsymbol{\mu}).
\)
Collecting the constant (determinant) terms yields the prefactor $C_\alpha^{(q)}$.
Collecting the quadratic and linear terms in $\boldsymbol{z}$ gives:
\[
-\tfrac12 \boldsymbol{z}^\top M_\alpha \boldsymbol{z} + h_\alpha^\top \boldsymbol{z} - K_\alpha,
\quad
M_\alpha=\boldsymbol{\Sigma}_q^{-1}+\alpha\,\boldsymbol{\Sigma}_j^{-1}-\alpha\,\boldsymbol{\Sigma}_{i^\star}^{-1},
\]
with $h_\alpha,K_\alpha$ as stated.
Completing the square and using the multivariate Gaussian integral
\(
\int \exp(-\tfrac12 \boldsymbol{z}^\top A \boldsymbol{z} + b^\top \boldsymbol{z})\,d\boldsymbol{z}=(2\pi)^{d/2}|A|^{-1/2}\exp(\tfrac12 b^\top A^{-1}b)
\)
(valid for $A\succ0$), and noticing that $(2\pi)^{d/2}$ cancels with the corresponding factor in $q$, we obtain the claimed closed form.

If $q=p_{i^\star}$ (i.e., $\boldsymbol{\mu}_q=\boldsymbol{\mu}_{i^\star}$ and $\boldsymbol{\Sigma}_q=\boldsymbol{\Sigma}_{i^\star}$), then we have:
\[
\mathbb E_{\boldsymbol{z}\sim p_{i^\star}}\!\left[\Big(\tfrac{p_j(\boldsymbol{z})}{p_{i^\star}(\boldsymbol{z})}\Big)^{\!\alpha}\right]
=\int p_{i^\star}(\boldsymbol{z})^{1-\alpha}p_j(\boldsymbol{z})^{\alpha}\,d\boldsymbol{z}
=\rho_\alpha\!\big(p_{i^\star},p_j\big),
\]
where the right-hand side is the standard multivariate Gaussian Chernoff $\alpha$-coefficient:
\[
\rho_{\alpha}(p_{i^\star},p_j)
= \frac{|\boldsymbol{\Sigma}_j|^{\alpha/2}\,|\boldsymbol{\Sigma}_{i^\star}|^{(1-\alpha)/2}}
     {\big|\alpha\,\boldsymbol{\Sigma}_j + (1-\alpha)\,\boldsymbol{\Sigma}_{i^\star}\big|^{1/2}}\,
  \exp\!\left(-\frac{\alpha(1-\alpha)}{2}\,
              (\boldsymbol{\mu}_j - \boldsymbol{\mu}_{i^\star})^\top
              \big(\alpha\,\boldsymbol{\Sigma}_j + (1-\alpha)\,\boldsymbol{\Sigma}_{i^\star}\big)^{-1}
              (\boldsymbol{\mu}_j - \boldsymbol{\mu}_{i^\star})
       \right),
\]
as given in \citet[Eq.~(35)]{nielsen2014generalized}.

Let {\small $N_2\ =\ \sum_{j\ne i^\star}\mathbf 1_{p_j \ge p_{i^\star}}.$} denote the number of rivals that beat {\small $i^\star$}.
Then the Top-$k$ error event under {\small $\boldsymbol{z}\sim q$} is {\small $\{N_q\ge k\}$}. Similar to the proof of Theorem~\ref{thm:topk}, we have:

\begin{align*}
    \Pr(N_2\ge k\mid \boldsymbol{z} \sim q) &\ \le\ \frac{\mathbb E[N_2\mid \boldsymbol{z} \sim q]}{k}    \\
    &\ = \ \frac{1}{k}\sum_{j\ne i^\star}\Pr(p_j(\boldsymbol{z})\ge  p_{i^\star}(\boldsymbol{z})\mid \boldsymbol{z} \sim q]) \\
&\ \le\ \frac{1}{k}\sum_{j\ne i^\star}\mathbb E_q\!\left[\Big(\tfrac{p_j(\boldsymbol{z})}{p_{i^\star}(\boldsymbol{z})}\Big)^{\!\alpha}\right], \\
&\ = \frac{1}{k}\sum_{j\ne i^\star} \ C_\alpha^j\;\,|M_\alpha^j|^{-1/2}\,
\exp\!\left(\tfrac12\,(h_\alpha^j)^\top (M_\alpha^j)^{-1} h_\alpha^j \,-\, K_\alpha^j\right)
\end{align*}

\section{Experimental Supplementary Material}\label{sec_app_exp}

\subsection{Details of Evaluation datasets and metrics}\label{subsec_dataset}
We employ a subset of the FLAN-v2 datasets~\citep{wei2022finetuned} for domain generation. 
FLAN-v2 datasets is a large-scale instruction-tuning corpus that integrates diverse Natural Language Understanding (NLU) and Natural Language Generation (NLG) tasks into an instruction–response format. 
A detailed summary of the selected datasets together with their associated evaluation metrics is provided below.

\textbf{\textit{Natural Language Inference}.} Natural language inference tasks require models to determine logical relations (entailment, contradiction, or neutrality) between pairs of sentences. We use the following datasets: (1) ANLI (v1-v3); (2) CB; (3) MNLI (matched, mismatched); (4) QNLI; (5) RTE; (6) SNLI; (7) WNLI. All datasets in this cluster are evaluated using accuracy as the metric.

\textbf{\textit{Question Answering}.}  
Question answering tasks evaluate the ability to retrieve or generate correct answers from passages or knowledge bases.  
We use the following datasets: (1) ARC (Challenge, Easy); (2) BoolQ; (3) MultiRC; (4) NaturalQuestions; (5) OpenBookQA; (6) ReCoRD; (7) SQuAD (v1-v2); (8) TriviaQA. For ARC, BoolQ, OpenBookQA, and ReCoRD, accuracy is used as the evaluation metric. For the remaining datasets, both accuracy and F1 score are reported.

\textbf{\textit{Sentiment Analysis}.}  
Sentiment analysis tasks involve classifying the polarity or emotional tone of text, such as positive or negative sentiment.  
We use the following datasets: (1) Sentiment140; (2) SST2.  
For Sentiment140, both accuracy and F1 score are reported, while SST2 is evaluated using accuracy only.

\textbf{\textit{Translation}.}  
Translation tasks test the capacity to generate fluent and semantically correct text across different languages.  
We use the following datasets: (1) ParaCrawl\_EnEs; (2) WMT14\_EnFr; (3) WMT16\_CsEn; (4) WMT16\_DeEn; (5) WMT16\_FiEn; (6) WMT16\_RoEn; (7) WMT16\_RuEn; (8) WMT16\_TrEn. All translation tasks are evaluated using BLEU, which measures $n$-gram overlap between system outputs and reference translations.

\textbf{\textit{Commonsense Reasoning}.}  
Commonsense reasoning tasks require leveraging everyday knowledge and logical inference to choose or generate plausible answers.  
We use the following datasets: (1) COPA; (2) HellaSwag; (3) PIQA; (4) StoryCloze. All datasets in this cluster are evaluated using accuracy as the metric.

\textbf{\textit{Paraphrase}.}  
Paraphrase tasks assess whether two sentences express the same underlying meaning, despite differences in wording.  
We use the following datasets: (1) GLUE\_MRPC; (2) GLUE\_QQP; (3) STS-B; (4) PAWS\_Wiki.  MRPC and QQP are evaluated with both accuracy and F1, while STS-B and PAWS\_Wiki are evaluated using accuracy.

\textbf{\textit{Struct-to-Text Generation}.}  
These tasks focus on converting structured data, such as triples or tables, into coherent natural language text.  
We use the following datasets: (1) CommonGen; (2) DART; (3) E2E\_NLG; (4) Gigaword; (5) WebNLG\_En. All datasets in this cluster are evaluated using ROUGE (ROUGE-1,2,L) and BLEU, since $n$-gram overlap captures the informativeness and fluency of generated text.   

\textbf{\textit{Coreference Resolution}.}  
Coreference tasks require identifying expressions in text that refer to the same entity.  
We use the following datasets: (1) Definite Pronoun Resolution; (2) WSC. Both datasets are evaluated using accuracy.  

\textbf{\textit{Text Correction}.}  
Text correction tasks involve detecting and fixing grammatical errors or inconsistencies in sentences.  
We use the following datasets: (1) CoLA; (2) FixPunct; (3) TrueCase.  All datasets in this cluster are evaluated using accuracy. 

\textbf{\textit{Word-level Tasks}.}  
Word-level tasks examine lexical semantics and basic text processing such as contextual meaning and segmentation.  
We use the following datasets: (1) WiC; (2) Word\_Segment.  WiC is evaluated using accuracy, while Word\_Segment is evaluated using both accuracy and F1.

Accuracy is sufficient when tasks have clear-cut, single-label predictions, such as classification or multiple-choice settings, where each prediction is either entirely correct or incorrect. In contrast, tasks with span-based, multi-label, or imbalanced data distributions may yield partially correct predictions. In these cases, F1 score is reported alongside accuracy, as it balances precision and recall and provides a more sensitive evaluation of partial correctness.

\subsection{Extended Visualization of LoRA Projections}
\label{sub_app_observation}
\begin{figure}[!htb]
    	\vspace{-4mm}
    	\centering
    	\includegraphics[width=1\textwidth]{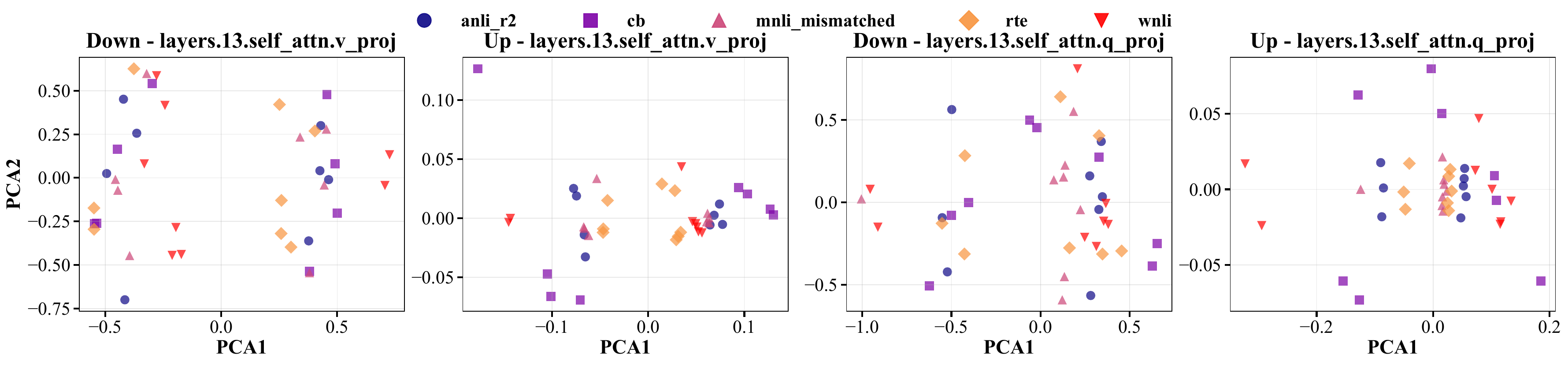}
    	\vspace{-8mm}
    	\caption{Scatter plots of the first two principal components derived from vectors in LoRA query and value projection matrices in layer~13 across five NLI tasks.}
    	\label{fig:within_13}
    	\vspace{-3mm}
\end{figure}

\begin{figure}[!htb]
    	\centering
    	\includegraphics[width=1\textwidth]{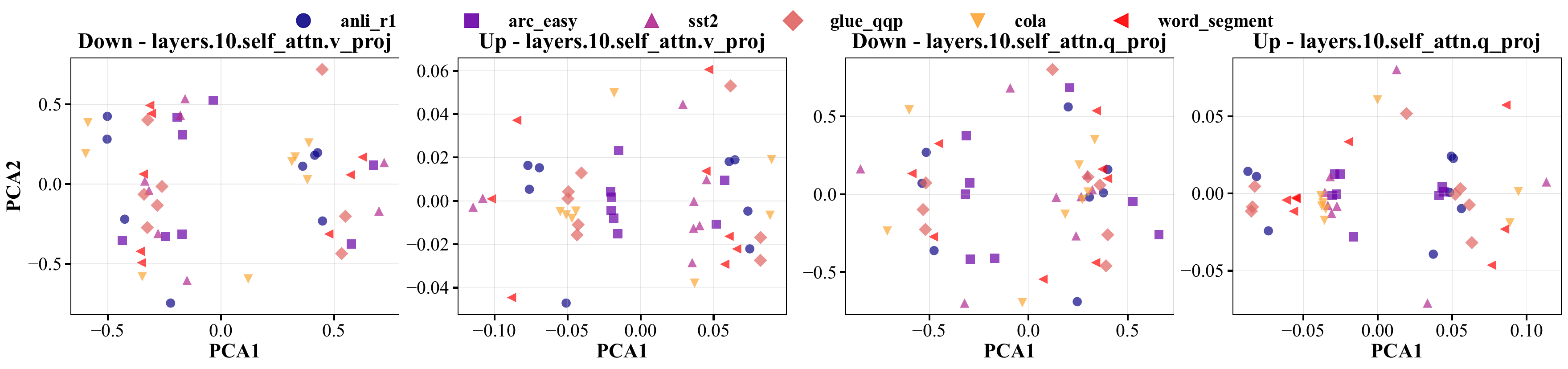}
        \includegraphics[width=1\textwidth]{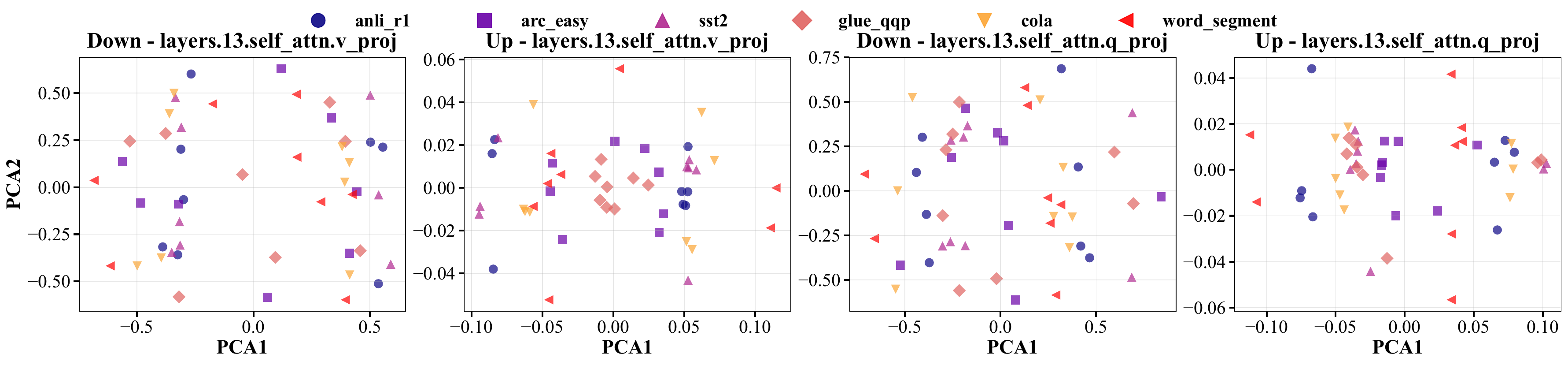}
    	\vspace{-8mm}
    	\caption{Scatter plots of the first two principal components derived from vectors in LoRA query and value projection matrices in layers~10 and 13 across six tasks from different clusters.}
    	\label{fig:cross}
    	\vspace{-3mm}
\end{figure}

In the main text, we reported scatter plots of the first two principal components obtained from vectors in LoRA projection matrices fine-tuned on five NLI tasks, focusing on two representative layers. To further validate these findings, Fig.~\ref{fig:within_13} presents results from additional layers, which reveal consistent structural patterns across model depth. We additionally extend the analysis to tasks drawn from different clusters (As shown in Fig.~\ref{fig:cross}), where similar trends are observed. Taken together, these results provide stronger empirical support for the key observations discussed in Sec.~\ref{subsection:observation}.

\subsection{Full Experimental Results} \label{subsec:result}
Tab.~\ref{tab:within_cluster_t5} reports the per-task accuracy of different methods on the NLI cluster using FLAN-T5-large.  Comprehensive performance under the unseen cluster setting is reported for all tasks, including detailed metrics for each task and evaluation criterion. Results for LLaMA2-7B are shown in Tab.\ref{tab:cross_cluster_llama}, and results for FLAN-T5-large are shown in Tab.\ref{tab:cross_cluster_t5}.
Fig.~\ref{fig:mapping} shows a heatmap of cosine similarities produced by \texttt{Retriever} across tasks, with tasks from the same cluster grouped by green boxes.

\begin{table}[htbp]
\centering
\caption{Detailed performance on the NLI cluster using FLAN-T5-large. Tasks with a white background are set as \textit{seen} tasks, while those with a gray background are set as \textit{unseen} tasks.
For each task, the best accuracy among all methods is in \textbf{bold}, and the second best is \underline{underlined}.}
\label{tab:within_cluster_t5}
\begin{adjustbox}{max width=\textwidth}
\begin{tabular}{l | c | c c c c c c c c c}
\toprule
\textbf{Methods} & \textbf{LoRA} & \textbf{HiLoRA} & \textbf{HiLoRA-GS} & \textbf{HiLoRA-ROC} & \textbf{Retriever} & \textbf{LEGO} & \textbf{Arrow} & \textbf{Phatgoose} & \textbf{Ensemble} & \textbf{Merged} \\
\midrule
\multicolumn{11}{c}{\textit{FLAN-T5-Large}}\\
\midrule
ANLI-r1    & 60.20 & 60.40 & \underline{61.20} & 57.90 & 60.70 & 60.60 & 60.80 & \textbf{61.30} & 60.70 & 60.80\\
\rowcolor{gray!20}
\cellcolor{white}ANLI-r2    & \cellcolor{white}43.30 & 42.20 & 42.70 & 41.00 & 42.80 & 42.50 & \textbf{43.70} & 42.90 & \underline{43.50} & 43.40  \\
ANLI-r3    & 44.50 & 43.08 & 44.42 & 42.42 & 44.25 & 44.17 & \textbf{45.25} & \underline{44.67} & 44.33 & 44.33\\
\rowcolor{gray!20}
\cellcolor{white}CB         & \cellcolor{white}78.00 & 78.00 & 78.00 & 78.00 & 78.00 & 78.00 & \textbf{80.00} & 78.00 & 78.00 & 78.00  \\
MNLI       & 88.59 & \textbf{89.12} & 84.77 & \underline{89.00} & 88.16 & 64.69 & 61.76 & 66.52 & 63.52 & 58.87\\
\rowcolor{gray!20}
\cellcolor{white}MNLI-mis   & \cellcolor{white}89.14 & \textbf{88.91} & 82.07 & \underline{88.52} & 88.67 & 62.93 & 68.79 & 64.10 & 61.05 & 56.76 \\
QNLI       & 82.54 & \textbf{82.70} & 82.54 & \textbf{82.70} & \textbf{82.70} & 82.38 & 81.95 & 81.02 & 82.38 & 80.66 \\
\rowcolor{gray!20}
\cellcolor{white}RTE        & \cellcolor{white}78.89 & 61.15 & 61.11 & 61.48 & 62.59 & 62.96 & \textbf{63.70} & 60.00 & \textbf{63.70} & 57.04 \\
SNLI       & 60.08 & \textbf{80.00} & 60.31 & 80.00 & 68.44 & 13.75 & 17.85 & 10.14 & 13.28 & 6.13\\
\rowcolor{gray!20}
\cellcolor{white}WNLI       & \cellcolor{white}52.86 & 51.43 & \underline{51.43} & 44.29 & 51.29 & 50.00 & \textbf{54.29} & 44.29 & 51.43 & 44.29  \\
\midrule
Avg        & 67.81 & \textbf{67.70} & 64.85 & 66.53 & \underline{66.76} & 56.20 & 57.81 & 55.29 & 56.19 & 53.03 \\
\bottomrule
\end{tabular}
\end{adjustbox}
\end{table}

\begin{table}[htbp]
\centering
\caption{Per-task performance of LLaMA2-7B under the cross-cluster setting. }
\label{tab:cross_cluster_llama}
\begin{adjustbox}{max width=\textwidth}
\centering
\begin{tabular}{l | c | c| c c c c c c c c c}
\toprule
\textbf{Tasks/Methods} & \textbf{Metric} &\textbf{LoRA} & \textbf{HiLoRA} & \textbf{HiLoRA-GS} & \textbf{HiLoRA-ROC} & \textbf{Retriever} & \textbf{LEGO} & \textbf{Arrow} & \textbf{Phatgoose} & \textbf{Ensemble} & \textbf{Merged} \\
\midrule
ANLI\_r1 & ACC & 46.40 & 30.70 & 27.30 & 27.70 & 30.00 & 28.20 & 31.30 & 29.40 & 29.90 & 21.40 \\
ANLI\_r2 & ACC &  40.10 & 34.50 & 32.50 & 30.20 & 34.30 & 30.30 & 32.20 & 31.50 & 33.00 & 22.00 \\
ANLI\_r3 &  ACC & 36.92 & 31.67 & 30.42 & 30.67 & 31.33 & 29.42 & 30.58 & 30.08 & 29.75 & 16.17 \\
CB & ACC &  80.00 & 70.00 & 62.00 & 72.00 & 66.00 & 60.00 & 52.00 & 64.00 & 64.00 & 4.00 \\
MNLI & ACC &  77.66 & 50.74 & 49.65 & 49.92 & 48.95 & 49.34 & 45.12 & 47.85 & 45.35 & 0.94 \\
MNLI\_mis & ACC &  79.69 & 51.29 & 49.45 & 50.59 & 49.45 & 50.82 & 45.66 & 48.05 & 46.37 & 1.21 \\
QNLI & ACC &  77.27 & 46.84 & 42.62 & 46.56 & 42.23 & 41.41 & 47.97 & 44.80 & 47.19 & 8.59 \\
RTE & ACC &  52.96 & 62.22 & 57.41 & 52.96 & 72.59 & 50.74 & 58.15 & 55.93 & 56.30 & 39.26 \\
SNLI & ACC &  67.42 & 40.31 & 41.09 & 40.23 & 15.82 & 40.08 & 34.18 & 39.06 & 36.68 & 0.51 \\
WNLI & ACC &  72.86 & 47.14 & 49.86 & 49.14 & 47.14 & 48.57 & 45.71 & 47.14 & 47.14 & 2.86 \\
\midrule
AVG\_NLI &  &  63.13 & 46.54 & 44.23 & 45.00 & 43.78 & 42.89 & 42.29 & 43.78 & 43.57 & 11.69 \\
\midrule
ARC\_C& ACC  & 40.43 & 34.74 & 30.78 & 30.43 & 34.31 & 32.33 & 30.00 & 32.76 & 31.90 & 0.43 \\
ARC\_E& ACC  & 40.17 & 48.64 & 44.28 & 45.89 & 47.50 & 46.74 & 41.57 & 46.06 & 45.42 & 0.55 \\
Bool\_Q & ACC & 85.23 & 78.36 & 75.86 & 64.34 & 77.27 & 75.43 & 64.45 & 76.41 & 76.09 & 21.21 \\
MultiRC & \makecell{ACC \\ F1} & \makecell{62.15 \\ 65.86} & \makecell{36.48 \\ 38.21} & \makecell{34.49 \\ 36.10} & \makecell{40.82 \\ 42.99} & \makecell{33.36 \\ 34.93} & \makecell{39.57 \\ 41.28} & \makecell{32.58 \\ 34.94} & \makecell{34.41 \\ 36.58} & \makecell{35.08 \\ 36.84} & \makecell{3.52 \\ 13.15} \\
NaturalQuestions &\makecell{ACC \\ F1}& \makecell{18.95 \\ 29.82} & \makecell{12.03 \\ 20.39} & \makecell{9.38 \\ 18.45} & \makecell{5.90 \\ 11.33} & \makecell{8.59 \\ 17.13} & \makecell{13.40 \\ 21.42} & \makecell{9.88 \\ 17.82} & \makecell{13.48 \\ 21.37} & \makecell{11.80 \\ 19.65} & \makecell{0.35 \\ 9.13} \\
OpenBookQA & ACC & 58.20 & 45.60 & 44.60 & 45.40 & 38.60 & 45.80 & 43.00 & 42.80 & 45.20 & 0.20 \\
RecoRD & ACC & 92.87 & 69.81 & 63.51 & 66.45 & 72.30 & 68.19 & 66.26 & 65.18 & 62.78 & 35.42 \\
SQuAD\_v1 &\makecell{ACC \\ F1}& \makecell{55.20 \\ 74.91} & \makecell{44.65 \\ 64.25} & \makecell{40.90 \\ 59.47} & \makecell{39.80 \\ 58.99} & \makecell{40.35 \\ 60.01} & \makecell{45.78 \\ 63.75} & \makecell{25.51 \\ 41.89} & \makecell{43.40 \\ 61.98} & \makecell{42.97 \\ 62.36} & \makecell{2.42 \\ 21.43} \\
SQuAD\_v2 & \makecell{ACC \\ F1} & \makecell{64.34 \\ 73.80} & \makecell{26.45 \\ 36.03} & \makecell{24.57 \\ 34.01} & \makecell{24.41 \\ 35.56} & \makecell{19.26 \\ 30.42} & \makecell{26.48 \\ 36.51} & \makecell{14.77 \\ 23.35} & \makecell{24.06 \\ 33.99} & \makecell{25.78 \\ 35.16} & \makecell{0.35 \\ 9.06} \\
TriviaQA & \makecell{ACC \\ F1} & \makecell{54.02 \\ 60.27} & \makecell{47.19 \\ 58.96} & \makecell{41.13 \\ 54.73} & \makecell{43.95 \\ 54.95} & \makecell{37.27 \\ 49.81} & \makecell{49.18 \\ 59.14} & \makecell{42.77 \\ 53.35} & \makecell{47.46 \\ 58.79} & \makecell{46.99 \\ 58.33} & \makecell{3.91 \\ 22.88} \\
\midrule
AVG\_QA & & 59.66 & 46.95 & 43.56 & 43.19 & 43.55 & 46.67 & 39.37 & 45.10 & 44.89 & 10.09 \\
\midrule
Sentiment140 & \makecell{ACC \\ F1} & \makecell{43.06 \\ 44.70} & \makecell{43.27 \\ 44.26} & \makecell{40.61 \\ 41.91} & \makecell{44.49 \\ 45.51} & \makecell{39.59 \\ 41.34} & \makecell{42.73 \\ 43.68} & \makecell{31.84 \\ 33.05} & \makecell{42.22 \\ 43.67} & \makecell{40.41 \\ 41.08} & \makecell{4.49 \\ 12.05} \\
SST2 & ACC & 75.86 & 63.10 & 58.51 & 62.99 & 59.77 & 62.64 & 49.08 & 63.06 & 59.77 & 0.11 \\
\midrule
AVG\_Sentiment && 59.87 & 54.43 & 49.88 & 54.00 & 50.12 & 52.93 & 40.76 & 53.00 & 50.26 & 4.19 \\
\midrule
ParaCrawl\_EnEs& BLUE  & 29.05 & 27.18 & 28.61 & 18.15 & 15.25 & 19.31 & 27.01 & 26.11 & 25.34 & 11.12 \\
WMT14\_EnFr & BLUE & 30.49 & 30.23 & 30.95 & 23.20 & 12.57 & 25.79 & 29.67 & 29.83 & 29.91 & 15.64 \\
WMT16\_CsEn & BLUE & 19.67 & 18.56 & 19.37 & 12.55 & 7.52 & 14.10 & 18.26 & 18.04 & 18.29 & 11.76 \\
WMT16\_DeEn & BLUE & 26.72 & 27.26 & 27.40 & 20.06 & 11.22 & 21.67 & 26.57 & 26.19 & 26.61 & 16.12 \\
WMT16\_FiEn & BLUE & 14.58 & 14.63 & 15.36 & 10.31 & 5.40 & 11.20 & 14.58 & 14.23 & 14.50 & 9.04 \\
WMT16\_RoEn & BLUE & 24.91 & 22.80 & 22.87 & 16.19 & 12.01 & 17.57 & 22.47 & 22.13 & 22.33 & 13.72 \\
WMT16\_RuEn & BLUE & 22.27 & 17.84 & 21.54 & 13.33 & 10.08 & 16.82 & 20.99 & 19.46 & 21.31 & 12.50 \\
WMT16\_TrEn & BLUE & 8.11 & 7.74 & 8.33 & 5.54 & 1.91 & 5.12 & 7.93 & 7.79 & 7.88 & 5.39 \\
\midrule
AVG\_Translation && 21.98 & 20.78 & 21.80 & 14.92 & 9.50 & 16.45 & 20.93 & 20.47 & 20.77 & 11.91 \\
\midrule
COPA & ACC & 72.00 & 65.00 & 66.00 & 66.00 & 71.00 & 68.00 & 59.00 & 62.00 & 63.00 & 20.00 \\
HellaSwag & ACC& 71.76 & 28.87 & 22.42 & 26.41 & 22.85 & 23.55 & 28.98 & 23.91 & 26.99 & 0.00 \\
PIQA & ACC& 61.75 & 53.72 & 50.98 & 48.58 & 53.22 & 47.70 & 51.75 & 52.90 & 51.64 & 1.80 \\
StoryCloze & ACC& 62.94 & 63.48 & 61.66 & 64.17 & 32.89 & 61.28 & 63.58 & 64.71 & 66.47 & 39.14 \\
\midrule
AVG\_Commonsense& & 67.11 & 52.76 & 50.27 & 51.29 & 44.99 & 50.14 & 50.83 & 50.88 & 52.03 & 15.24 \\
\midrule
GLUE\_MRPC & \makecell{ACC \\ F1} & \makecell{68.00 \\ 68.00} & \makecell{65.25 \\ 65.25} & \makecell{65.75 \\ 65.75} & \makecell{53.50 \\ 53.50} & \makecell{58.50 \\ 58.50} & \makecell{36.75 \\ 36.75} & \makecell{53.25 \\ 53.25} & \makecell{59.50 \\ 59.50} & \makecell{58.75 \\ 58.75} & \makecell{17.50 \\ 19.50} \\
GLUE\_QQP & \makecell{ACC \\ F1} & \makecell{76.13 \\ 76.13} & \makecell{64.80 \\ 64.80} & \makecell{67.66 \\ 67.66} & \makecell{53.44 \\ 53.44} & \makecell{68.12 \\ 68.12} & \makecell{58.71 \\ 58.71} & \makecell{63.09 \\ 63.09} & \makecell{56.80 \\ 56.80} & \makecell{64.73 \\ 64.73} & \makecell{2.23 \\ 3.01} \\
STSB & ACC& 34.82 & 17.83 & 20.33 & 16.78 & 19.78 & 15.95 & 17.27 & 16.78 & 16.43 & 0.21 \\
PAWS\_Wiki & ACC& 88.55 & 64.45 & 46.68 & 47.19 & 71.64 & 48.24 & 46.76 & 56.17 & 56.33 & 9.10 \\
\midrule
AVG\_Paraphrase && 66.88 & 53.08 & 50.11 & 42.73 & 54.51 & 39.91 & 45.09 & 47.31 & 49.06 & 7.61 \\
\midrule
CommonGen & \makecell{ROUGE-1 \\ ROUGE-2 \\ROUGE-L \\ BLEU}& \makecell{54.60 \\ 23.18 \\ 47.82 \\ 11.95} & \makecell{43.27 \\ 14.05 \\ 36.84 \\ 6.74} & \makecell{44.89 \\ 13.88 \\ 36.30 \\ 7.21} & \makecell{43.08 \\ 2.21 \\ 28.31 \\ 0.27} & \makecell{43.03 \\ 8.65 \\ 32.51 \\ 3.90} & \makecell{24.14 \\ 1.47 \\ 21.70 \\ 0.08} & \makecell{39.70 \\ 10.48 \\ 32.57 \\ 4.20} & \makecell{38.95 \\ 12.26 \\ 33.79 \\ 5.60} & \makecell{37.43 \\ 11.26 \\ 32.47 \\ 5.08} & \makecell{35.44 \\ 9.68 \\ 27.93 \\ 3.71} \\
DART & \makecell{ROUGE-1 \\ ROUGE-2 \\ROUGE-L \\ BLEU}& \makecell{72.16 \\ 47.78 \\ 56.07 \\ 36.44} & \makecell{52.74 \\ 25.66 \\ 39.76 \\ 14.99} & \makecell{52.89 \\ 25.48 \\ 39.68 \\ 14.60} & \makecell{51.78 \\ 23.79 \\ 38.74 \\ 13.19} & \makecell{53.03 \\ 24.49 \\ 39.44 \\ 13.20} & \makecell{35.26 \\ 16.03 \\ 28.32 \\ 6.39} & \makecell{49.97 \\ 26.05 \\ 39.27 \\ 14.55} & \makecell{50.33 \\ 26.08 \\ 39.19 \\ 15.74} & \makecell{50.61 \\ 26.19 \\ 39.45 \\ 15.26} & \makecell{42.92 \\ 21.06 \\ 33.72 \\ 10.43} \\
E2E\_NLG & \makecell{ROUGE-1 \\ ROUGE-2 \\ROUGE-L \\ BLEU}& \makecell{73.04 \\ 44.84 \\ 52.67 \\ 31.79} & \makecell{56.60 \\ 29.79 \\ 41.28 \\ 17.99} & \makecell{58.55 \\ 29.29 \\ 41.45 \\ 16.56} & \makecell{47.01 \\ 21.20 \\ 33.49 \\ 7.75} & \makecell{51.66 \\ 27.65 \\ 39.45 \\ 13.89} & \makecell{18.73 \\ 9.24 \\ 16.72 \\ 0.28} & \makecell{59.00 \\ 33.25 \\ 43.66 \\ 21.17} & \makecell{57.95 \\ 32.76 \\ 43.81 \\ 18.94} & \makecell{56.15 \\ 31.78 \\ 42.31 \\ 18.24} & \makecell{54.62 \\ 29.99 \\ 40.50 \\ 20.17} \\
Gigaword & \makecell{ROUGE-1 \\ ROUGE-2 \\ROUGE-L \\ BLEU}& \makecell{36.28 \\ 16.07 \\ 32.90 \\ 10.32} & \makecell{25.28 \\ 8.58 \\ 21.67 \\ 3.30} & \makecell{25.24 \\ 8.57 \\ 21.70 \\ 3.37} & \makecell{22.26 \\ 7.46 \\ 19.11 \\ 2.82} & \makecell{25.18 \\ 8.31 \\ 21.63 \\ 3.08} & \makecell{24.64 \\ 8.27 \\ 21.21 \\ 3.42} & \makecell{25.43 \\ 8.64 \\ 21.90 \\ 3.37} & \makecell{26.50 \\ 9.50 \\ 22.90 \\ 4.09} & \makecell{26.31 \\ 9.55 \\ 22.90 \\ 4.03} & \makecell{22.92 \\ 7.76 \\ 19.54 \\ 3.19} \\
WebNLG\_En & \makecell{ROUGE-1 \\ ROUGE-2 \\ROUGE-L \\ BLEU}& \makecell{78.37 \\ 56.19 \\ 62.52 \\ 45.18} & \makecell{48.01 \\ 24.95 \\ 39.14 \\ 15.60} & \makecell{48.29 \\ 25.53 \\ 39.16 \\ 15.02} & \makecell{52.02 \\ 25.93 \\ 42.23 \\ 14.60} & \makecell{53.28 \\ 26.66 \\ 42.80 \\ 14.62} & \makecell{32.22 \\ 15.57 \\ 28.15 \\ 5.92} & \makecell{45.27 \\ 24.09 \\ 37.65 \\ 13.94} & \makecell{45.01 \\ 24.60 \\ 37.52 \\ 14.73} & \makecell{42.99 \\ 22.90 \\ 35.71 \\ 13.65} & \makecell{43.94 \\ 23.76 \\ 35.89 \\ 11.58} \\
\midrule
AVG\_Text\_Generation & & 44.51 & 28.31 & 28.18 & 24.86 & 27.32 & 15.89 & 27.71 & 28.01 & 27.21 & 24.94 \\
\midrule
DPR &ACC& 45.89 & 60.18 & 61.07 & 56.61 & 58.04 & 58.57 & 61.07 & 59.46 & 58.39 & 6.96 \\
WSC & ACC& 50.00 & 63.00 & 63.00 & 62.00 & 60.00 & 59.00 & 61.00 & 57.00 & 63.00 & 7.00 \\
\midrule
avg\_coreference && 47.95 & 61.59 & 62.04 & 59.30 & 59.02 & 58.79 & 61.04 & 58.23 & 60.70 & 6.98 \\
\midrule
CoLA &ACC& 62.24 & 55.49 & 55.30 & 53.18 & 54.82 & 55.78 & 55.78 & 55.20 & 55.30 & 5.20 \\
FixPunct & ACC & 34.69 & 22.85 & 22.85 & 17.93 & 21.09 & 15.94 & 21.13 & 21.09 & 21.17 & 6.60 \\
TrueCase & ACC & 67.27 & 14.61 & 21.48 & 6.09 & 2.50 & 0.39 & 11.13 & 14.45 & 13.32 & 7.23 \\
\midrule
AVG\_Text\_Correct & & 54.73 & 30.98 & 33.21 & 25.73 & 26.14 & 24.04 & 29.35 & 29.58 & 29.93 & 6.34 \\
\midrule
WIC & ACC& 57.78 & 49.68 & 49.68 & 49.68 & 50.63 & 49.37 & 51.49 & 50.32 & 48.95 & 0.16 \\
Word\_Segment & \makecell{ACC \\ F1}& \makecell{62.46 \\ 90.07} & \makecell{25.70 \\ 60.04} & \makecell{23.44 \\ 59.24} & \makecell{18.87 \\ 54.07} & \makecell{19.57 \\ 66.07} & \makecell{14.57 \\ 41.13} & \makecell{24.07 \\ 55.88} & \makecell{24.06 \\ 57.03} & \makecell{23.78 \\ 58.67} & \makecell{10.74 \\ 34.80} \\
\midrule
AVG\_Word && 67.02 & 46.28 & 45.51 & 43.08 & 46.73 & 38.61 & 45.73 & 45.43 & 43.09 & 11.47 \\
\bottomrule
\end{tabular}
\end{adjustbox}
\end{table}

\begin{table}[htbp]
\centering
\caption{Per-task performance of FLAN-T5-large under the cross-cluster setting. }
\label{tab:cross_cluster_t5}
\begin{adjustbox}{max width=\textwidth}
\centering
\begin{tabular}{l | c | c| c c c c c c c c c}
\toprule
\textbf{Tasks/Methods} & \textbf{Metric} &\textbf{LoRA} & \textbf{HiLoRA} & \textbf{HiLoRA-GS} & \textbf{HiLoRA-ROC} & \textbf{Retriever} & \textbf{LEGO} & \textbf{Arrow} & \textbf{Phatgoose} & \textbf{Ensemble} & \textbf{Merged} \\
\midrule
ANLI\_r1 & ACC & 60.20 & 57.90 & 59.20 & 54.00 & 60.60 & 60.70 & 60.50 & 60.00 & 60.60 & 60.60 \\
ANLI\_r2 & ACC & 43.30 & 42.70 & 42.20 & 41.40 & 42.00 & 42.80 & 43.10 & 42.90 & 43.00 & 42.70 \\
ANLI\_r3 & ACC & 44.50 & 43.42 & 44.50 & 41.83 & 43.83 & 44.00 & 44.00 & 44.25 & 43.92 & 44.00 \\
CB & ACC & 78.00 & 78.00 & 78.00 & 80.00 & 80.00 & 78.00 & 78.00 & 82.00 & 78.00 & 78.00 \\
MNLI & ACC & 88.59 & 81.76 & 82.50 & 83.48 & 83.40 & 58.24 & 56.52 & 79.49 & 59.10 & 50.78 \\
MNLI\_mis & ACC & 89.14 & 82.66 & 83.40 & 83.87 & 84.77 & 55.62 & 50.20 & 79.10 & 56.37 & 48.20 \\
QNLI & ACC & 82.54 & 81.09 & 80.78 & 82.38 & 82.23 & 77.93 & 75.78 & 80.98 & 78.44 & 74.10 \\
RTE & ACC & 78.89 & 74.07 & 67.04 & 74.81 & 73.33 & 59.26 & 53.33 & 72.96 & 60.00 & 52.59 \\
SNLI & ACC & 60.08 & 36.13 & 10.35 & 38.87 & 10.27 & 3.79 & 1.64 & 23.40 & 3.83 & 1.52 \\
WNLI & ACC & 52.86 & 57.14 & 38.57 & 51.43 & 60.00 & 41.43 & 42.86 & 55.71 & 44.29 & 38.57 \\
\midrule
AVG\_NLI &  & 67.81 & 63.49 & 58.65 & 63.21 & 62.04 & 52.18 & 50.59 & 62.08 & 52.75 & 49.11 \\
\midrule
Bool\_Q & ACC & 87.27 & 79.22 & 76.09 & 81.60 & 80.70 & 74.73 & 73.05 & 80.04 & 75.23 & 70.39 \\
MultiRC & \makecell{ACC \\ F1} & \makecell{55.31 \\ 58.71} & \makecell{54.02 \\ 57.19} & \makecell{53.48 \\ 56.76} & \makecell{52.42 \\ 55.64} & \makecell{53.20 \\ 56.53} & \makecell{52.97 \\ 56.30} & \makecell{53.59 \\ 56.94} & \makecell{54.02 \\ 57.12} & \makecell{53.24 \\ 56.53} & \makecell{52.77 \\ 56.11} \\
SQuAD\_v1 & \makecell{ACC \\ F1} & \makecell{42.66 \\ 64.25} & \makecell{36.21 \\ 59.92} & \makecell{34.30 \\ 57.50} & \makecell{36.33 \\ 57.56} & \makecell{27.38 \\ 51.23} & \makecell{30.27 \\ 53.92} & \makecell{29.45 \\ 53.01} & \makecell{35.00 \\ 58.28} & \makecell{30.94 \\ 54.31} & \makecell{29.92 \\ 53.29} \\
SQuAD\_v2 & \makecell{ACC \\ F1} & \makecell{66.56 \\ 77.13} & \makecell{65.39 \\ 76.39} & \makecell{64.26 \\ 75.36} & \makecell{64.34 \\ 75.16} & \makecell{62.97 \\ 74.24} & \makecell{63.09 \\ 74.27} & \makecell{62.46 \\ 73.67} & \makecell{64.77 \\ 75.78} & \makecell{63.09 \\ 74.53} & \makecell{61.91 \\ 73.30} \\
\midrule
AVG\_QA &  & 67.39 & 63.44 & 61.73 & 63.08 & 60.87 & 60.03 & 59.40 & 63.13 & 60.39 & 58.51 \\
\midrule
Sentiment140 & \makecell{ACC \\ F1} & \makecell{42.04 \\ 43.18} & \makecell{41.84 \\ 43.41} & \makecell{41.63 \\ 43.10} & \makecell{41.63 \\ 43.21} & \makecell{41.63 \\ 43.09} & \makecell{41.63 \\ 43.01} & \makecell{41.63 \\ 43.10} & \makecell{41.43 \\ 42.82} & \makecell{41.63 \\ 43.02} & \makecell{41.63 \\ 43.02} \\
SST2 & ACC & 75.75 & 74.48 & 73.91 & 76.55 & 73.10 & 73.91 & 73.56 & 74.14 & 73.68 & 73.56 \\
\midrule
AVG\_Sentiment &  & 59.18 & 58.55 & 58.14 & 58.49 & 57.73 & 58.11 & 57.96 & 58.13 & 58.00 & 57.94 \\
\midrule
ParaCrawl\_EnEs & BLEU & 27.15 & 26.78 & 26.80 & 26.30 & 26.93 & 26.71 & 26.61 & 26.66 & 26.72 & 26.53 \\
WMT16\_RoEn & BLEU & 20.81 & 20.79 & 20.93 & 20.51 & 20.83 & 20.87 & 20.97 & 20.79 & 20.88 & 20.88 \\
WMT16\_TrEn & BLEU & 8.94 & 8.81 & 8.66 & 8.84 & 8.86 & 8.74 & 8.65 & 8.37 & 8.71 & 8.55 \\
\midrule
AVG\_Translation &  & 18.97 & 18.79 & 18.80 & 18.55 & 18.88 & 18.77 & 18.74 & 18.61 & 18.77 & 18.65 \\
\midrule
GLUE\_MRPC & \makecell{ACC \\ F1} & \makecell{89.00 \\ 89.00} & \makecell{81.75 \\ 81.75} & \makecell{82.00 \\ 82.00} & \makecell{76.50 \\ 76.50} & \makecell{77.00 \\ 77.00} & \makecell{80.75 \\ 80.75} & \makecell{80.25 \\ 80.25} & \makecell{81.50 \\ 81.50} & \makecell{81.00 \\ 81.00} & \makecell{81.25 \\ 81.25} \\
GLUE\_QQP & \makecell{ACC \\ F1} & \makecell{85.43 \\ 85.43} & \makecell{83.36 \\ 83.40} & \makecell{82.42 \\ 82.89} & \makecell{79.45 \\ 79.45} & \makecell{82.19 \\ 82.19} & \makecell{78.40 \\ 81.41} & \makecell{76.02 \\ 80.16} & \makecell{82.19 \\ 82.19} & \makecell{78.32 \\ 81.60} & \makecell{70.78 \\ 80.51} \\
STSB & ACC & 44.29 & 41.99 & 41.23 & 31.89 & 37.95 & 40.39 & 39.42 & 41.57 & 41.30 & 41.02 \\
PAQS\_Wiki & ACC & 94.61 & 93.59 & 93.75 & 84.14 & 92.93 & 93.48 & 93.63 & 93.79 & 93.63 & 93.95 \\
\midrule
AVG\_Paraphrase & & 78.33 & 75.18 & 74.91 & 68.00 & 72.52 & 73.63 & 72.85 & 74.76 & 73.97 & 72.96 \\
\midrule
DART & \makecell{ROUGE-1 \\ ROUGE-2 \\ ROUGE-L \\ BLEU} & \makecell{76.03 \\ 54.63 \\ 61.99 \\ 46.56} & \makecell{75.82 \\ 54.39 \\ 61.76 \\ 46.66} & \makecell{75.75 \\ 54.22 \\ 61.52 \\ 46.73} & \makecell{75.95 \\ 54.27 \\ 61.61 \\ 45.78} & \makecell{75.85 \\ 54.38 \\ 61.78 \\ 46.57} & \makecell{75.77 \\ 54.21 \\ 61.62 \\ 46.65} & \makecell{75.79 \\ 54.32 \\ 61.66 \\ 46.76} & \makecell{75.66 \\ 54.17 \\ 61.50 \\ 46.53} & \makecell{75.72 \\ 54.19 \\ 61.48 \\ 46.68} & \makecell{75.60 \\ 54.11 \\ 61.43 \\ 46.72} \\
E2E\_NLG & \makecell{ROUGE-1 \\ ROUGE-2 \\ ROUGE-L \\ BLEU} & \makecell{73.08 \\ 46.57 \\ 54.16 \\ 35.27} & \makecell{73.48 \\ 46.73 \\ 54.21 \\ 35.43} & \makecell{73.27 \\ 46.62 \\ 54.04 \\ 35.37} & \makecell{73.95 \\ 46.92 \\ 54.40 \\ 35.02} & \makecell{73.55 \\ 46.77 \\ 54.25 \\ 35.44} & \makecell{73.23 \\ 46.63 \\ 54.15 \\ 35.33} & \makecell{73.09 \\ 46.47 \\ 53.96 \\ 35.12} & \makecell{73.29 \\ 46.66 \\ 54.12 \\ 35.38} & \makecell{73.19 \\ 46.57 \\ 54.04 \\ 35.28} & \makecell{73.14 \\ 46.53 \\ 53.97 \\ 35.23} \\
WebNLG\_En & \makecell{ROUGE-1 \\ ROUGE-2 \\ ROUGE-L \\ BLEU} & \makecell{83.34 \\ 64.51 \\ 69.18 \\ 56.89} & \makecell{82.50 \\ 63.24 \\ 68.12 \\ 55.83} & \makecell{82.71 \\ 63.39 \\ 68.19 \\ 56.18} & \makecell{81.77 \\ 62.52 \\ 67.77 \\ 53.07} & \makecell{82.61 \\ 63.35 \\ 68.25 \\ 55.72} & \makecell{82.70 \\ 63.29 \\ 68.04 \\ 55.98} & \makecell{82.78 \\ 63.35 \\ 67.98 \\ 56.26} & \makecell{82.49 \\ 63.26 \\ 68.02 \\ 56.11} & \makecell{82.75 \\ 63.36 \\ 68.03 \\ 56.16} & \makecell{82.71 \\ 63.27 \\ 67.96 \\ 56.36} \\
\midrule
AVG\_Text\_Generation &  & 60.18 & 59.85 & 59.83 & 59.42 & 59.88 & 59.80 & 59.79 & 59.76 & 59.79 & 59.75 \\
\midrule
DPR & ACC & 86.25 & 76.79 & 76.25 & 78.21 & 76.07 & 75.89 & 75.89 & 77.14 & 76.07 & 75.36 \\
WSC & ACC & 40.00 & 51.00 & 47.00 & 57.00 & 48.00 & 46.00 & 46.00 & 47.00 & 48.00 & 46.00 \\
\midrule
AVG\_Coreference & & 63.12 & 63.89 & 61.62 & 63.61 & 62.04 & 60.95 & 60.95 & 62.07 & 62.04 & 60.68 \\
\midrule
CoLA & ACC & 64.26 & 58.00 & 56.55 & 59.25 & 56.17 & 56.94 & 56.36 & 56.84 & 56.94 & 56.45 \\
FixPunct & ACC & 41.25 & 39.45 & 40.16 & 37.27 & 39.61 & 40.47 & 40.55 & 40.39 & 40.62 & 40.62 \\
TrueCase & ACC & 59.22 & 67.03 & 65.94 & 64.53 & 66.25 & 66.29 & 66.45 & 66.80 & 66.33 & 65.55 \\
\midrule
AVG\_Text\_Correct &  & 54.91 & 54.83 & 54.21 & 53.68 & 54.01 & 54.56 & 54.45 & 54.68 & 54.63 & 54.21 \\
\midrule
WIC & ACC & 66.98 & 65.71 & 65.40 & 48.25 & 65.40 & 66.03 & 63.81 & 64.92 & 66.19 & 65.40 \\
Word\_Segment & \makecell{ACC \\ F1} & \makecell{63.91 \\ 88.33} & \makecell{67.15 \\ 90.83} & \makecell{67.34 \\ 90.74} & \makecell{66.72 \\ 92.82} & \makecell{67.07 \\ 90.52} & \makecell{70.62 \\ 92.76} & \makecell{70.20 \\ 92.56} & \makecell{71.53 \\ 93.15} & \makecell{70.55 \\ 92.72} & \makecell{70.35 \\ 92.47} \\
\midrule
AVG\_Word & & 71.55 & 73.35 & 72.22 & 64.01 & 72.10 & 73.86 & 72.59 & 73.63 & 73.91 & 73.40 \\
\bottomrule
\end{tabular}
\end{adjustbox}
\end{table}

\begin{figure}[htb!]
    \centering
    \includegraphics[width=0.4\linewidth]{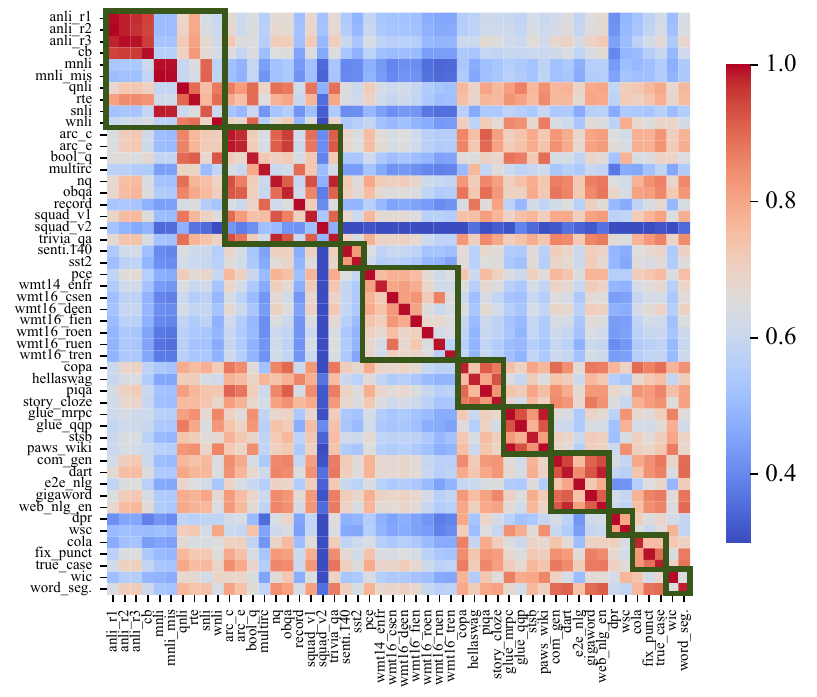}
    \vspace{-5mm}
    \caption{Input-LoRA similarity heatmap produced by \texttt{Retriever}, where tasks from the same cluster are enclosed within green boxes for clarity.}
    \label{fig:mapping_retri}
\end{figure}

\section{LLM Usage}
Large Language Models (LLMs) were used solely to aid in the writing and polishing of the manuscript.
LLMs, specifically ChatGPT, were employed exclusively as writing assistants in the preparation of this manuscript. Their role was limited to improving the presentation quality of the text, including tasks such as rephrasing sentences, correcting grammar, enhancing readability, and improving the overall flow of exposition. The use of LLMs was confined to linguistic refinement, and they were not involved in generating, verifying, or shaping any scientific ideas.

All research contributions, including the formulation of research questions, algorithmic design, theoretical derivations, and experimental studies, were conceived and executed entirely by the authors.  
Their contribution was restricted to stylistic and grammatical adjustments, with no bearing on the substance of the research.

The authors retain full responsibility for the entire content of this work, including any text improved with LLM assistance. We have carefully ensured that the usage of LLMs complies with ethical standards and does not introduce plagiarism, fabrication, or other forms of scientific misconduct.

\end{document}